\documentclass{article}

\usepackage{geometry}
\usepackage{framed,multirow}
\usepackage{amssymb}
\usepackage{latexsym}

\usepackage{url}
\usepackage[table]{xcolor}
\usepackage{soul}
\definecolor{newcolor}{rgb}{.8,.349,.1}
\definecolor{darkred}{rgb}{1.,.4,.4}
\definecolor{darkgreen}{rgb}{.4,1.,.4}
\definecolor{lightred}{rgb}{1.,.8,.8}
\definecolor{lightgreen}{rgb}{.8,1.,.8}
\definecolor{modified}{rgb}{.3,.3,.9}
\definecolor{modified}{rgb}{0.,0.,0.}
\usepackage{amsmath}
\usepackage{bm}
\usepackage[all]{xy}
\usepackage{subcaption}
\usepackage{adjustbox}
\usepackage{pgfplots}
\usepackage{algorithm}
\usepackage{algorithmic}

\usepackage{enumerate}
\usepackage{makecell}

\usepackage{amsthm}
\newtheorem{thm}{Theorem}[section]

\newtheorem{Thm}[thm]{Theorem}
\newtheorem{Prop}[thm]{Proposition}
\newtheorem{Lem}[thm]{Lemma}
\newtheorem{Corl}[thm]{Corollary}

\theoremstyle{definition}
\newtheorem{Def}[thm]{Definition}

\theoremstyle{remark}
\newtheorem*{Rem}{Remark}

\newcommand{\mM}{\mathcal{M}}
\newcommand{\mH}{\mathcal{H}}

\newcommand{\mX}{\mathcal{X}}
\newcommand{\mY}{\mathcal{Y}}
\newcommand{\mZ}{\mathcal{Z}}

\newcommand{\mE}{\mathcal{E}}
\newcommand{\mD}{\mathcal{D}}
\newcommand{\mS}{\mathcal{S}}
\newcommand{\mG}{\mathcal{G}}
\newcommand{\mL}{\mathcal{L}}

\newcommand{\md}{\mathrm{d}}
\newcommand{\mR}{\mathbb{R}}
\newcommand{\trans}{\mathsf{T}}
\newcommand{\me}{\mathrm{e}}
\newcommand{\mi}{\mathrm{i}}
\DeclareMathOperator*{\argmin}{argmin}
\DeclareMathOperator*{\diag}{diag}

\DeclareMathOperator{\Span}{Span}

\usepackage{hyperref}
\hypersetup{
	colorlinks=true,
	linkcolor=blue,
	filecolor=magenta,
	urlcolor=red,
}
\usepackage{authblk}
\providecommand{\keywords}[1]{\textbf{\textit{Keywords---}} #1}
\title{Latent assimilation with implicit neural representations\\ for unknown dynamics}
\author[1]{\small Zhuoyuan Li\thanks{Corresponding author. e-mail: zy.li@stu.pku.edu.cn}}
\author[2,3]{\small Bin Dong}
\author[4,1]{\small Pingwen Zhang}

\affil[1]{\footnotesize School of Mathematical Sciences, Peking University, Beijing 100871, China}
\affil[2]{\footnotesize Beijing International Center for Mathematical Research, Peking University, Beijing 100871, China}
\affil[3]{\footnotesize Center for Machine Learning Research, Peking University, Beijing 100871, China}
\affil[4]{\footnotesize School of Mathematics and Statistics, Wuhan University, Wuhan 430072, China}
\date{}
\begin{document}
\maketitle
\begin{abstract}
	Data assimilation is crucial in a wide range of applications, but it often faces challenges such as high computational costs due to data dimensionality and incomplete understanding of underlying mechanisms. To address these challenges, this study presents a novel assimilation framework, termed Latent Assimilation with Implicit Neural Representations (LAINR). By introducing Spherical Implicit Neural Representations (SINR) along with a data-driven uncertainty estimator of the trained neural networks, LAINR enhances efficiency in the assimilation process. Experimental results indicate that LAINR holds a certain advantage over existing methods based on AutoEncoders, both in terms of accuracy and efficiency.
	Besides, the flexibility of LAINR makes it a potential solution for real-world applications.
\end{abstract}

\section{Introduction}
Data assimilation (DA) has increasingly become an essential tool across a multitude of disciplines, including but not limited to weather forecasting, climate modeling, oceanography, ecology, and even economics \cite{Kalnay2002,DA4ClimateResearch,RoleOfDA4Ecology,DA4Ecology,DA4Economics}. It aims to integrate information from various data sources, such as in-situ measurements, satellite, and radar observations \cite{Lahoz2014}, with existing physical models that describe a system's underlying dynamics. The objective of DA is to obtain the best estimate of the system states while accounting for uncertainties in both observations and physical models.

From a mathematical standpoint, DA offers an efficient and practical solution to a specific type of optimal control problem \cite{Bocquet2016} 
\begin{equation}\label{eq:original-continuous-DA}
	\begin{cases}
		\dot{\bm u}(t)=\mM(\bm u(t),t)+\bm\varepsilon^M(t), & 0\le t\le T, \\
		\bm u(0)=\bm u_0
	\end{cases}
\end{equation}
with the objective function
\begin{equation}
	\bm J(\bm u_0)=\mL_B\left(\bm u_0,\bm u^b\right)+\int_0^T\mL_O\left(\mH(\bm u(t)),\bm y(t),t\right)\md t+\int_0^T\mL_E\left(\bm\varepsilon^M(t)\right)\md t
\end{equation}
to be minimized. Here, the system dynamics is denoted by $\mM$, and $\mH$ stands for the observation operator. $\mL_B$ and $\mL_O$ penalize the discrepancy between the system states and the background estimate $\bm u^b$ and the observation data $\bm y(t)$, respectively, while $\mL_E$ penalizes the model error $\bm\varepsilon^M(t)$. The goal is to find the optimal control $\bm u_0$ as the initial system state that minimizes the objective function $\bm J(\bm u_0)$. For instance, in the realm of modern numerical weather prediction (NWP), $\bm u(t)$ may consist of the meteorological variables of interest governed by atmospherical dynamics, then DA frameworks are employed to determine the optimal system states for further prediction. In practice, dealing with the optimization problem \eqref{eq:original-continuous-DA} defined on an infinite-dimensional functional space is challenging. Therefore, the problem is typically reformulated by discretizing on both the spatial and temporal dimensions, which leads to minimizing
\begin{equation}\label{eq:discretized-DA-physical}
	\bm J(\bm x_0)=\mL_B\left(\bm x_0,\bm x^b\right)+\sum_k\mL_O\left(\mH_k(\bm x_k),\bm y_k\right)+\sum_k\mL_E\left(\bm\varepsilon_k^M\right)
\end{equation}
subject to
\begin{equation}\label{eq:discretized-DA-physical-constraints}
	\bm x_{k+1}=\mM_k(\bm x_k)+\bm\varepsilon_k^M,\quad k=0,1,\cdots,
\end{equation}
where the subscripts $k$ denote temporal indices.
\textcolor{modified}{Here, $\bm x=\{\bm u(\bm p)\}_{\bm p\in S}$ represents a discrete sampling of $\bm u$ with a sampling set $S$.
To model the discrete dynamics,
a numerical approximation $\mM_k$ for integrating $\mM(\cdot,t)$ in \eqref{eq:original-continuous-DA} is typically employed, which inherently encompasses the integration steps required for transitioning between successive time steps. Similarly, the accumulated error $\bm\varepsilon_k^M$ is obtained from $\bm\varepsilon^M(t)$ in \eqref{eq:original-continuous-DA}.
Besides, $\mH_k$ stands for the observation operator discretized on the sampling set.}
In this work, we refer to such downsampled version $\bm x$ as the ``physical state'' in contrast to the ``latent state'' $\bm z$ to be introduced later, and one should not confuse the physical state $\bm x$ with the system state $\bm u$ since the physical state does not contain all the physical features of the system state in general.

Early efforts for DA primarily revolved around techniques like nudging and optimal (statistical) interpolation methods. Subsequently, variational methods (3/4D-Var) have been proposed based on the optimal control theory. A key advantage of the variational approaches lies in their ability to make meteorological fields comply with the dynamical equations of NWP models, while simultaneously minimizing the discrepancies between simulations and observations. However, these methods are often computationally intensive, particularly for high-dimensional systems. Additionally, the necessity to derive the corresponding tangent linear models and adjoint models becomes another challenge, especially for complex dynamics.

Another advanced development of DA frameworks is the proposal of Kalman Filter (KF) methods \cite{Geir2009,Bocquet2016}, which facilitate sequential assimilation of data, and thus offer benefits for real-time applications. Literature such as \cite{Talagrand2014} has shown that Kalman filtering is equivalent to variational assimilation in some limits. Unfortunately, the challenge of high dimensionality persists in that the scales of the computational grid and the underlying matrices are linked to the state dimension. Some approximate or suboptimal Kalman filters have been developed to partially alleviate the issue, but at the expense of confining the assimilated states within a low-dimensional subspace, potentially harming the overall performance.

Moreover, existing DA frameworks rely on accurate dynamical cores. Nevertheless, capturing small-scale physical processes or edge effects is still a challenge, especially where the physical knowledge is limited and no well-designed parameterizations are available, which will probably lead to significant and undetermined model errors. Another notable obstacle arises when dealing with highly non-linear systems. The inherent difficulties in optimizing non-convex problems and the inconsistency with the linear assumptions of the Kalman filter point to a need for more advanced DA techniques.

One approach to handle the high-dimensionality issue is Reduced-Order-Models (ROMs), which establish a lower-dimensional parameterization space and then shift the DA framework to the latent space.
Classical ROMs usually employ linear trial subspaces for model reduction, and they require the subspace of dimension much larger than the intrinsic dimension of the solution manifold to ensure high accuracy, especially for highly non-linear problems \cite{Ohlberger2015ReducedBasis}. Based on the Kolmogorov $n$-width \cite{Pinkus1985}, \cite{DeVore1989optimal} has proposed a continuous non-linear version
\begin{equation}\label{eq:continuous-non-linear-n-width}
	d_n(\mS)=\inf_{\substack{P:\mS\to\mR^n\\Q:\mR^n\to\mS}}\sup_{f\in\mS}\left\|f-Q(P(f))\right\|
\end{equation}
to quantify the intrinsic complexity of a compact functional space $K$, which together with the recent work \cite{Lee2020,Fresca2021DL-ROM,ye2023analysis} shows the theoretical superiority of non-linear ROMs over linear ones.

Machine Learning (ML), and specifically Deep Learning (DL) has emerged as an effective data-driven tool to approximate high-dimensional mappings thanks to its strengths to extract intricate structures in high-dimensional data \cite{lecun2015dlnature}, and consequently, neural networks are used not only to establish non-linear ROMs, whose training processes depend on reconstruction loss functions similar to \eqref{eq:continuous-non-linear-n-width} where the mappings $P$ and $Q$ are translated as parameterized encoders and decoders, but also to learn the corresponding reduced or latent dynamics \cite{gruber2022comparison,Regazzoni2019,chen2023crom} from the encoded data. Provided with a well-trained encoder-decoder pair denoted by $\mE_{\bm\theta}$ and $\mD_{\bm\varphi}$, respectively, we may obtain a latent state $\bm z_k$ for each $\bm x_k$ from \eqref{eq:discretized-DA-physical}\eqref{eq:discretized-DA-physical-constraints} such that
\begin{equation}
	\bm z_k=\mE_{\bm\theta}(\bm x_k),\,\bm x_k=\mD_{\bm\varphi}(\bm z_k)+\bm\varepsilon_k^{\textrm{E-D}}
\end{equation}
with minor reconstruction error $\bm\varepsilon_k^{\textrm{E-D}}$. Furthermore, if we additionally succeed in attaining a latent \textcolor{modified}{transition} $\mG_{\bm\psi,k}$ such that
\begin{equation}
	\bm z_{k+1}=\mG_{\bm\psi,k}(\bm z_k)+\bm\varepsilon^L_k
\end{equation}
with minor prediction error $\bm\varepsilon^L_k$, we may solve the assimilation problem by minimizing the objective function
\begin{equation}
	\bm J(\bm z_0)=\mL_B\left(\bm z_0,\bm z^b\right)+\sum_k\mL_O\left(\mH_k\left(\mD_{\bm \varphi}(\bm x_k)\right),\bm y_k\right),
\end{equation}
on a lower-dimensional space where all the latent states $\{\bm z_k\}_k$ reside.
Such a transition for DA from physical space to latent space is referred to as the latent assimilation (LA) \cite{Amendola2021CAE-DA}. The latent dynamical model $\mG_{\bm\psi,k}$ is usually data-driven, and no expert knowledge is required. By optimizing on a lower-dimensional space, LA has the benefit of reducing both the storage and computational demands compared with traditional methods.

In recent years, the rapid development of DL has had a transformative impact on various domains, including computer vision, natural language processing, and more recently scientific computing. The applications of DL in solving complex physical problems, such as solving differential equations \cite{PINNs,DeepRitz,DeepGalerkin}, discovering underlying physical laws \cite{PDENet2.0,SINDy,AI-Feyman}, and simulating spatio-temporal dynamics \cite{DeepONet,FNO}, have attracted significant attention and achieved promising results. In the specific area of atmospheric sciences, DL has proven itself highly effective in handling a variety of tasks, ranging from weather nowcasting and forecasting to post-processing and data assimilation.
Specifically, the success of forecast models like FourCastNet \cite{FourCastNet}, Pangu-Weather \cite{Pangu}, GraphCast \cite{GraphCast} and FengWu \cite{FengWu} focusing on learning surrogate dynamics has gradually dissolved the initial doubts \cite{Schultz2021CanDL} regarding DL's capability to handle complex dynamics.

\begin{figure}
	\centering
	\includegraphics[width=.8\textwidth]{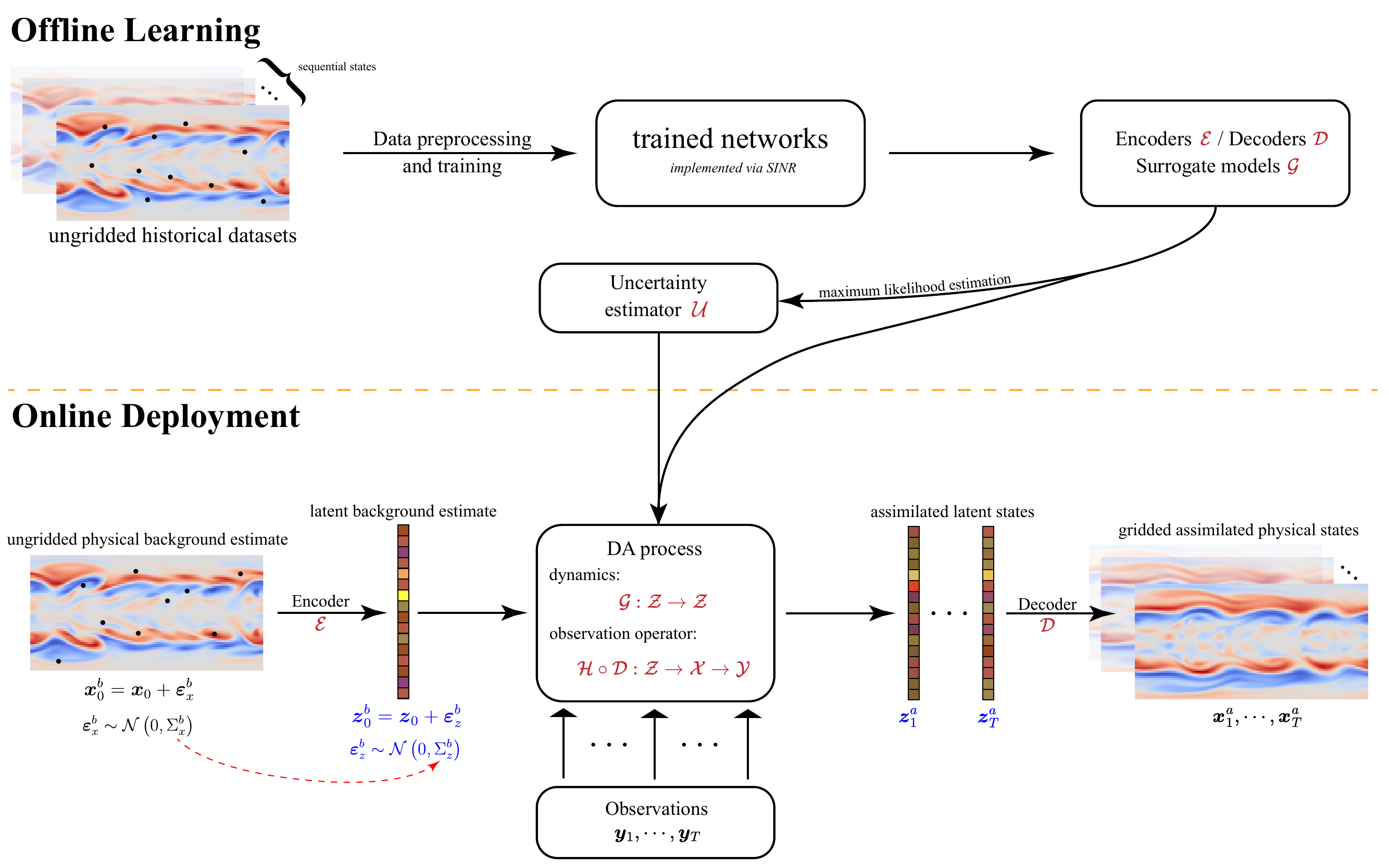}
	\caption{The pipeline of the LAINR framework.}
	\label{fig:LAINR-pipeline}
\end{figure}

In this work, we present the Latent Assimilation with Implicit Neural Representations (LAINR) framework (Figure \ref{fig:LAINR-pipeline}). Our framework is designed to assimilate data from systems with partially or even completely unknown dynamics. By proposing a new variant of the implicit neural representation, namely the Spherical Implicit Neural Representation (SINR), and a comprehensive assimilation procedure for learning surrogate models and estimating corresponding uncertainties, we aim to make a meaningful contribution to the ongoing exploration of DL and ML techniques in data assimilation and atmospheric sciences.


\section{Related works and contributions}
\paragraph{Reduced-order models}
The concept of Reduced-Order Models (ROMs) is not a new idea for data assimilation. As introduced previously, dimensionality reduction techniques are employed to represent high-dimensional vectors with lower-dimensional ones. Classical ROMs usually rely on linear methods like Proper Orthogonal Decomposition (POD) \cite{Cao2006,Artana2012,Arcucci2019}, Dynamic Mode Decomposition (DMD) \cite{Schmid2010} or wavelet decomposition \cite{Tangborn2000wavelet} to extract system's dominant modes from historical physical states, then the corresponding parameterizations $\bm z\in\mZ$ for each $\bm x\in\mX'\subseteq\mX$ are the projection coefficient vectors onto each mode.
Recently, DL advancements have offered alternative approaches to construct or enhance ROMs. AutoEncoders \cite{bank2023autoencoders,AutoencoderandItsVariants} are one of the most common methods to learn a non-linear, lower-dimensional representation of the data, and the learned representations can then be used to study and understand the behavior of the system at a lower computational cost \cite{Lee2020,Fresca2021DL-ROM}. Meanwhile, Recurrent Neural Networks (RNNs) \cite{lipton2015critical} including the Long Short-Term Memory (LSTM) networks \cite{LSTM} and the Gated Recurrent Unit (GRU) networks \cite{GRU} have also been employed to model time-dependent problems and capture the temporal correlations in the dataset.
\paragraph{Latent assimilation}
Latent assimilation has also benefited from the advent of DL, particularly in constructing non-linear encoder-decoder mappings for assimilation in latent space. For instance, \cite{Peyron2021LAwithAE} selects simple fully-connected deep neural networks for both the encoder and the decoder, with ReZero \cite{bachlechner2021rezero} as a variant of ResNet \cite{He2015ResNet} to enhance the stability of learning the latent dynamics. In spite of its effectiveness of dimensionality reduction, it is faced with scalability limitations since the fully-connected architecture can hardly be applied to multi-dimensional cases due to the unaffordable costs of storage and training for the huge number of network parameters. Recurrent Neural Networks (RNNs) can serve as an alternative way for merging the encoder-decoder structure and evolving latent dynamics simultaneously as proposed in \cite{LatentspaceDA-RNN}. The training of RNNs takes advantage of Reservoir Computing (RC) \cite{ReservoirComputing} to make the forward propagation faster, but as suggested in \cite{Arcomano2020MLGAFM,LatentspaceDA-RNN}, the RC approaches are more suitable for short-term prediction tasks. Meanwhile, creating patches when scaling up to high-dimensional problems will probably result in disagreement between neighborhoods. To effectively capture long-term dependencies in sequential data, LSTM architectures have been employed in \cite{ROM-DA,GeneralizedLA}, but they face the same issues as those of \cite{Peyron2021LAwithAE} that the model is difficult to scale up to handle high-dimensional problems. Finally, all the architectures mentioned above require the input data to lie on certain fixed grids, and the inevitable interpolation procedures probably bring additional errors.
Table \ref{tab:related_works} summarizes the comparative strengths and weaknesses of these methods as well as our proposed framework introduced later.
\renewcommand{\arraystretch}{1.2}
\begin{table}
	\centering
	\caption{Comparison of different methods for latent assimilation. The method names are borrowed from the original papers.}
	\begin{tabular}{c|c|c|c|c|c}
		\hline
		method                             & encoder-decoder  & latent dynamics & scalability & efficiency & flexibility \\
		\hline
		ETKF-Q-L \cite{Peyron2021LAwithAE} & fully-connected  & ReZero          & low         & high       & no          \\
		RNN-ETKF \cite{LatentspaceDA-RNN}  & RNN              & RC              & medium      & high       & no          \\
		NIROM-DA \cite{ROM-DA}             & POD/PCA          & LSTM            & low         & medium     & no          \\
		GLA \cite{GeneralizedLA}           & POD/PCA + Conv1d & LSTM            & low         & medium     & no          \\
		LAINR (ours)                       & SINR             & Neural ODE      & high        & medium     & yes         \\
		\hline
	\end{tabular}
	\label{tab:related_works}
\end{table}
\renewcommand{\arraystretch}{1.0}

\paragraph{Implicit neural representations}
Implicit Neural Representations (INRs) have emerged as an effective tool for encoding high-dimensional data, geometrical objects, or even functions in a compact and parameterized manner. Diverging from common neural networks representing data in the form of grids or graphs, INRs model them as continuous functions. Such approach has shown its superiority in a variety of applications, including shape modeling \cite{chen2019learning,park2019deepsdf}, image processing \cite{sitzmann2019siren,Bemana2020xfields,dupont2022coinpp}, and spatio-temporal dynamics \cite{Jiang2020MeshfreeFlowNet,chen2023crom,yin2023dino}. One key advantage of INRs lies in their agnosticism to data structures, leading to a high degree of flexibility. Moreover, similar to AutoEncoders, the non-linearity inherent to neural networks enables them to capture non-linear correlations, which have more potential than classical linear ROMs. In this work, we propose a new variant of INRs specially designed for 2D spherical data, namely the Spherical Implicit Neural Representations (SINRs), which aims to capture the complex dynamics of the system in a low-dimensional latent space based on the spherical harmonics.

The key aspects of our contributions can be summarized below.
\begin{itemize}
	\item \textbf{A Novel Mesh-free DA Framework:} We present a new assimilation framework named Latent Assimilation with Implicit Neural Representation (LAINR), which is mesh-free and applicable to multi-dimensional unknown dynamics.
	      \begin{itemize}
		      \item \underline{\textit{SINRs}.} We have formulated a specialized variant of INRs called the Spherical Implicit Neural Representations (SINRs) to handle 2D spherical data. SINRs are capable of capturing the system's complex dynamics at a lower cost by embedding the physical states into a low-dimensional latent space while ensuring high accuracy and reliable convergence guarantee.
		      \item \underline{\textit{Uncertainty Estimators}.} A simple but effective uncertainty qualification method has been developed to estimate uncertainties associated with both the latent surrogate model and the encoding process, leveraging maximum likelihood estimation (MLE) techniques.
	      \end{itemize}
	\item \textbf{Interoperability with Existing DA Algorithms:} LAINR has been engineered for compatibility to work seamlessly with existing data assimilation algorithms, thereby enhancing the performances and extending their applications. This feature encourages wider adoption for future advancements in the field of DA.
	\item \textbf{Experimental Validation:} We have conducted experiments on multi-dimensional cases, including an ideal shallow-water model and a real meteorological dataset (ERA5) to demonstrate the superiority of LAINR compared to existing AutoEncoder-based methods. The effectiveness of LAINR in capturing spatio-temporal dynamics, seamlessly integrating with existing DA algorithms and handling unstructured and unseen data is emphasized in the empirical findings.
\end{itemize}

The remainder of this paper is structured as follows. First, an overview of the existing LA approaches as well as their limitations are introduced in Section \ref{sec:latent-space-embedding-and-dynamics-learning}, which is followed by Section \ref{sec:LAINR} that presents our proposed LAINR framework in detail. Section \ref{sec:experiments} describes the experimental setup and exhibits the model performances. Finally, Section \ref{sec:conclusion} concludes the main features of our LAINR framework and outlines some of the potential avenues for future research.

\section{Latent assimilation}\label{sec:latent-space-embedding-and-dynamics-learning}
\begin{figure}[!h]
	\centering
	\begin{subfigure}{.45\textwidth}
		\[
			\xymatrix{
				\cdots\ar@[red][r]^\mM&\mX\ar@[red][r]^{\mM}\ar@[blue][d]_{\mH}&\mX\ar@[blue][d]^{\mH}\ar@[red][r]^\mM&\cdots\\
				&\mY&\mY&
			}
		\]
		\caption{classical DA that operates on the physical space}
		\label{fig:comp-DA}
	\end{subfigure}
	\hfill
	\begin{subfigure}{.45\textwidth}
		\[
			\xymatrix{
			\cdots\ar@[red][r]^\mG&\mZ\ar@[red][r]^{\mG}\ar@/^1ex/@[blue][d]^{\mD}&\mZ\ar@/^1ex/@[blue][d]^{\mD}\ar@[red][r]^\mG&\cdots\\
			\cdots\ar[r]^\mM&\mX\ar@/^1ex/[u]^{\mE}\ar[r]^{\mM}\ar@[blue][d]_{\mH}&\mX\ar@/^1ex/[u]^{\mE}\ar@[blue][d]^{\mH}\ar[r]^\mM&\cdots\\
			&\mY&\mY&
			}
		\]
		\caption{LA that operates on the latent space}
		\label{fig:comp-LA}
	\end{subfigure}
	\caption{Illustrations of classical DA and LA frameworks with colored forward operators (red) and observation operators (blue). The temporal indices here are omitted for simplicity.}
	\label{fig:comp-DA-and-LA}
\end{figure}
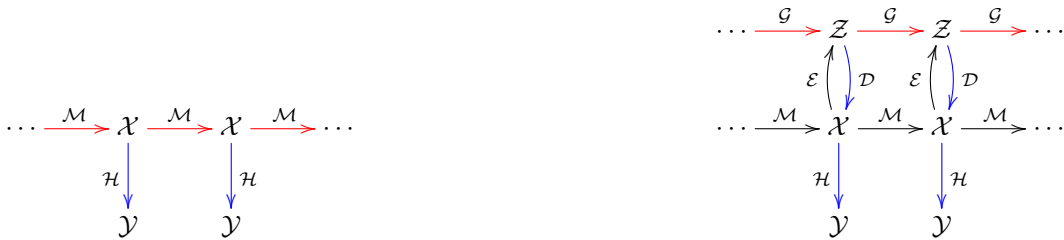

Latent Assimilation (LA) is essentially a specialized form of Data Assimilation (DA) methodology that makes use of dimensionality reduction of the assimilation space in order to save computational and storage costs. It characterizes the high-dimensional state (physical) space $\mX$, with a more compact, lower-dimensional parameterization (latent) space $\mZ$. By establishing a well-designed encoder-decoder pair capable of transforming data between the physical space and the latent space with high accuracy, LA shifts the physical states, physical dynamics, and the assimilation mechanism from $\mX$ onto $\mZ$. Same as conventional DA methodologies, LA also requires a forward propagation operator within the latent space $\mZ$ and an associated observation operator that links the latent space $\mZ$ to the observational space $\mY$ (Figure \ref{fig:comp-DA-and-LA}). Additionally, the uncertainties should also be accounted for within these mappings if possible.


\subsection{Embeddings from physical spaces to latent spaces}
LA framework explores suitable latent embeddings with the assumption that a low-dimensional parameterization exists for the high-dimensional physical states. Let $\mX=\mR^n$ and $\mZ=\mR^m$ be the physical state space and the latent space, respectively, where $n$ is considerably larger than $m$. The goal is to identify an encoder $\mE:\mX'\to\mZ$ and a decoder $\mD:\mZ\to\mX'$ that function as appropriate inverse mappings between the low-dimensional submanifold $\mX'\subseteq\mX$ encapsulating all possible physical states, and the latent space $\mZ$.
Here, we use the notation $\bm u:\mR^d\to\mR^c$ for the target system state to be assimilated, where $d$ signifies the spatial dimension of the dynamical system (usually $2$ or $3$), and $c$ refers to the number of features under consideration, such as horizontal wind, temperature, humidity, and so forth. For a fixed sampling set $S\subseteq\mR^d$, several strategies can be established to construct ROMs for the physical states or the snapshots $\bm x_k=\{\bm u_k(\bm p)\}_{\bm p\in S}\in\mR^{|S|\times c}=\mR^n=\mX$ corresponding to the system states $\bm u_k$. In this section, we review some prior approaches, including the classical linear ROMs and DL-based ROMs usually implemented via AutoEncoder structures.

\subsubsection{Classical linear ROMs}\label{sec:linear-encoders}
Linear Reduced-Order Models (ROMs) for representing system states have been extensively studied in the literature. Here, the word ``linear'' means that by fixing a collection of basis vectors $\{\bm\phi_j\}_j$ of the underlying functional space, the snapshot $\bm x$ for each fixed time step is approximated as a linear combination of the basis vectors
\begin{equation}
	\bm x\approx\sum_{j=1}^m c_j\bm\phi_j,
\end{equation}
where $c_j$ denotes the coefficient of the $j$-th basis vector and $m$ is the reduced dimension. In this context, the snapshot $\bm x$ is encoded as a lower-dimensional vector $\bm c=(c_1,\cdots,c_m)^\trans$. We only review some classical approaches here and choose not to dive deeply since they are not the focus of this work.

Proper Orthogonal Decomposition (POD), more specifically, the snapshot POD \cite{Lumley1970,LAWRENCE1987}, is a well-known procedure to extract an optimal basis from an ensemble of observations, which is determined by minimizing the empirical average of the associated projection error for each snapshot. For simplicity we introduce it in the context of scalar fields: given a collection of numerical and/or experimental observations $\{u_k:\mR^d\to\mR\}_{k=1}^K\subseteq L^2$ with temporal index $k=1,2\cdots,K$, POD solves the optimization problem
\begin{equation}\label{eq:pod-problem}
	\phi=\argmin_{\|\phi\|_{L^2}=1}\frac1K\sum_{k=1}^K\left|\left\langle u_k,\phi\right\rangle\right|^2,
\end{equation}
where $\langle\cdot,\cdot\rangle$ stands for the inner product on the $L^2$ space. A necessary condition for \eqref{eq:pod-problem} can be derived by calculus of variations, which leads to the following eigenvalue problem
\begin{equation}
	\frac1K\sum_{k=1}^Ku_k(s)\int u_k(s')\phi(s')\md s'=\lambda\phi(s),
\end{equation}
and it remains to use Fourier modes \cite{Berkooz1993} or apply eigenvalue decompositions to the correlation matrix of the snapshots in a discrete sense \cite{Cizmas2003} to derive the solution. Once the mapping $\phi$ is achieved, $\bm\phi_j$ is obtained by evaluating on the same sampling set $S\subseteq\mR^d$ for each solution $\phi$.

Dynamic Mode Decomposition (DMD) is another reduction approach based on the linearity assumption that
\begin{equation}\label{eq:dmd-assumption}
	\bm x_{k+1}=\bm A\bm x_k,\,k=0,1,\cdots,
\end{equation}
where $\bm A$ is a time-invariant linear operator. Such model reduction technique makes use of temporal correlations of the underlying dynamics, and the assumption \eqref{eq:dmd-assumption} amounts to a linear tangent approximation for non-linear dynamics \cite{Schmid2010}. By denoting $\bm B_1=(\bm x_0,\cdots,\bm x_{K-1})$ and $\bm B_2=(\bm x_1,\cdots,\bm x_K)$,
the singular-value decomposition (SVD) of $\bm B_1=\bm U\bm\Sigma\bm V^\trans$ helps to derive the dynamic modes $\bm\phi_j=\bm U\bm v_j$, where $\bm v_j$ is the $j$-th eigenvector of the matrix $\bm U^\trans\bm B_2\bm V\bm\Sigma^{-1}$.



For linear ROMs, each snapshot $\bm x$ is linearly parameterized in the latent space $\mZ$ as the coefficient vector $\bm c$. Taking POD as an example, the corresponding encoder and decoder can be formulated as
\begin{equation}\label{eq:POD-encoder-decoder}
	\mE^{\textrm{POD}}(\bm x)=\bm V\bm x,\quad\mD^{\textrm{POD}}(\bm z)=\bm V^\trans\bm z,
\end{equation}
where $\bm V$ has its rows being the eigenvectors or the so-called dominant modes of the correlation matrix. Such linear ROMs are not only straightforward to implement but also offer simple linear bijective mappings between $\mZ$ and $\mX'$, which simplify the derivation of the latent dynamics on $\mZ$ as long as the physical dynamics are provided. However, the reconstruction accuracy is heavily dependent on the number of basis functions, and as the system grows in complexity and non-linearity, many more basis functions are required to capture the small-scale structures, which diminishes the intended benefit of dimensionality reduction.
\subsubsection{AutoEncoders}\label{sec:autoencoders}
AutoEncoder (AE), consisting of an encoder and a decoder, is a common architecture in deep learning and serves as a viable non-linear method for dimensionality reduction. With the encoder and the decoder
\begin{equation}\label{eq:AE-encoder-decoder}
	\mE^{\textrm{AE}}(\bm x)=f_{\bm\theta}(\bm x),\quad\mD^{\textrm{AE}}(\bm z)=g_{\bm\varphi}(\bm z)
\end{equation}
implemented via neural networks, AutoEncoder is capable of extracting lower-dimensional latent representations from high-dimensional input data. Usually, the encoder progressively decreases the data dimension with its multi-layered structure, and by mirroring the encoder's architecture, the decoder takes the encoded representations and then reconstructs the original input data.

The objective of an AutoEncoder is to minimize the empirical reconstruction error by employing a loss function to measure the divergence between the input and the reconstructed data. The optimization
\begin{equation}
	\bm\theta,\bm\varphi=\argmin_{\bm\theta,\bm\varphi}\sum_{k=1}^K\mathcal{L}\left(\bm x_k,g_{\bm\varphi}(f_{\bm\theta}(\bm x_k))\right)
\end{equation}
promotes the learning of compressed meaningful data representation, where the function $\mathcal{L}$ measures the discrepancy between the inputs $\bm x_k$ and the reconstructed states $g_{\bm\varphi}(f_{\bm\theta}(\bm x_k))$. Once both the encoder $f_{\bm\theta}$ and the decoder $g_{\bm\varphi}$ are trained, $\bm z_k=f_{\bm\theta}(\bm x_k)$ is the corresponding latent representation of $\bm x_k$, and the decoder $g_{\bm\varphi}$ can be used to reconstruct the original data $\bm x_k$ from $\bm z_k$ conversely.

Compared with classical linear ROMs, AutoEncoders are capable of capturing non-linear relationships in the data, thereby allowing for the possibility of reducing complex dynamical system in physical space to simpler latent dynamics. 
Nonetheless, several limitations persist. Unlike classical linear ROMs, AutoEncoders do not offer an optimal representation since the minimizer is not guaranteed to be obtained. They may also encounter scalability challenges when dealing with high-dimensional problems as the number of grid points will increase exponentially concerning the state dimension and resolutions. Such complexity could lead to difficulties in designing effective structures. Moreover, AutoEncoders are generally restricted to operating on inputs with fixed structural attributes, limiting their utility in real-world applications that frequently involve partially missing data, ungridded data, or data with varying resolutions.
\subsection{Evolution of latent dynamics}\label{sec:evolution}
Establishing a suitable surrogate model for the evolution of latent dynamics forms another crucial part of the LA framework, which not only accounts for the system's inherent non-linearity but also incorporates the auto-regressive nature of the latent dynamics evolution, regardless of whether or not the explicit form of the physical dynamics is provided. We simplify our discussion by assuming time-invariant latent dynamics, which is approximated by a parameterized surrogate model $\mG_{\bm\psi}$ such that
\begin{equation}
	\bm z_k\mapsto\mG_{\bm\psi}(\bm z_k)\approx\bm z_{k+1},\, k=0,1,\cdots.
\end{equation}
The goal is to minimize the discrepancies between subsequent latent states, $\bm z_{k+1}$ and $\mG_{\bm\psi}(\bm z_{k})$. While linear regression serves as a straightforward method for constructing the latent surrogate model, its performance is limited in describing highly non-linear dynamics. As an alternative, the neural network offers a common data-driven approach to capturing the non-linear relationships in the data and approximating the latent dynamics.

The choices for the structure of $\mG_{\bm\psi}$ can be quite diverse, depending on the specific characteristics of the system state's temporal dynamics. For instance, Long Short-Term Memory (LSTM) \cite{LSTM} and Gated Recurrent Units (GRUs) \cite{GRU} networks are specially designed for sequential data, whose structures enable the network to remember or forget past states based on the context, and the work of \cite{ROM-DA,GeneralizedLA} has followed such way. ResNet \cite{He2015ResNet} is another approach to modeling the sequential data, which coincides with the forward Euler method as a classical numerical integrator for solving differential equations. Embedding the physical dynamics into a latent space naturally leads to latent dynamics
\begin{equation}
	\frac{\md \bm z}{\md t}=\bm F(\bm z)
\end{equation}
with an unknown operator $\bm F$. The equation can then be discretized as
\begin{equation}
	\bm z_{k+1}\approx\bm z_k+\bm F(\bm z_k)\Delta t\approx\bm z_k+\bm f_{\bm \psi}(\bm z_k)\Delta t=:\mG_{\bm \psi}(\bm z_k),
\end{equation}
where a neural network $\bm f_{\bm\psi}$ is employed to approximate the unknown $\bm F$, and the trainable parameters are denoted by $\bm\psi$.
ResNet-based architectures have been adopted to learn the latent dynamics \cite{Peyron2021LAwithAE} via ReZero \cite{bachlechner2021rezero} to enhance the performance. However, such discretization methods are subject to temporal resolution and step size constraints, and applying them directly to real-world assimilation deployment with varing time steps may introduce additional errors.


\subsection{Assimilation in latent spaces}\label{sec:LA-framework}
Both the LA frameworks and classical DA frameworks require observed data $\{\bm y_k\}_k$, observation operators $\{\mH_k\}_k$ for each time step and an initial background estimate $\bm x_0$ for start-up. They also need certain noise modeling mechanism, which plays a crucial role in tracking error evolution and subsequently deriving the assimilated state. The main difference is that LA can learn the latent dynamics from a historical dataset, and thus more suitable when the physical dynamics are costly or unavailable. Once the encoder-decoder mappings $\mE$, $\mD$ and the surrogate model $\mG$ have been trained, a similar assimilation routine can be executed on the latent spaces. Next, we show the procedure of LA by taking the classical Ensemble Kalman Filter (EnKF) as an example.

First, we establish the encoder $\mE:\mX'\to\mZ$ and the decoder $\mD:\mZ\to\mX'$ according to equations such as \eqref{eq:POD-encoder-decoder}\eqref{eq:AE-encoder-decoder}. Concurrently, a surrogate model $\mG:\mZ\to\mZ$ gets trained as described in Section \ref{sec:evolution} such that
\begin{equation}
	\mD\circ\mE\approx\textrm{identity mapping on }\mX';\qquad\mD\circ\mG\circ\mE\approx\mM.
\end{equation}
In other words, the encoder $\mE$ and the decoder $\mD$ are nearly inverse mappings such that the diagram in Figure \ref{fig:comp-LA} commutes up to a small error. After that, the encoder gives the latent background estimate by defining $\bm z_0^b=\mE\left(\bm x_0^b\right)$, which is then used to generate ensemble members for the initial state $\left\{\bm z_0^{a,j}\right\}_j$. Here, we denote the ensemble index by $j$. Applying perturbed forward propagation in the latent space gives
\begin{equation}
	\bm z_k^{b,j}=\mG_k\left(\bm z_{k-1}^{a,j}\right)+\bm\varepsilon_k^{L,j},\quad\bm\varepsilon_k^{L,j}\sim\mathcal{N}\left(0,\bm\Sigma_k^L\right),
\end{equation}
where $\bm\Sigma_k^L$ indicates the covariance matrix of the error for the latent surrogate model $\mS_k$ at the $k$-th step. As for the analysis phase, the Kalman gain matrix can be calculated via
\begin{equation}
	\bm K_k=\bm P_k^b\bm H_k^\trans\left(\bm H_k\bm P_k^b\bm H_k^\trans+\bm\Sigma_k^o\right)^{-1}.
\end{equation}
The matrix $\bm P_k^b$ is the covariance matrix for the background ensembles $\left\{\bm z_k^{b,j}\right\}_j$, and the matrix $\bm H_k$ is the tangent linear model corresponding with the latent observation operator $\mH_k'=\mH_k\circ\mD$ mapping from the latent space to the observation space. Finally, the analyzed ensembles can be updated via
\begin{equation}
	\bm z_k^{a,j}=\bm z_k^{b,j}+\bm K_k\bm d_k^j,
\end{equation}
where the innovation vector for the $j$-th ensemble is computed as
\begin{equation}
	\bm d_k^j=\bm y_k^o+\bm\varepsilon_k^{o,j}-\mH_k'\left(\bm z_k^{b,j}\right),\quad\bm\varepsilon_k^{o,j}\sim\mathcal{N}\left(0,\bm\Sigma_k^o\right),
\end{equation}
with the observation data and the covariance matrix denoted by $\bm y_k^o$ and $\bm\Sigma_k^o$, respectively. A simple decoding mapping
\begin{equation}
	\bm x_k^a=\mD\left(\bm z_k^a\right)
\end{equation}
on the ensemble mean $\bm z_k^a$ can help obtain the analyzed physical state $\bm x_k^a$ whenever needed.

\section{LAINR framework}\label{sec:LAINR}
The central contribution of our work lies in the introduction of the Latent Assimilation with Implicit Neural Representations (LAINR) framework, an extension of the previously proposed LA framework. In LAINR, we borrow the idea of Implicit Neural Representations (INRs) to take advantage of its flexibility. More precisely, we establish the Spherical Implicit Neural Representations (SINRs) utilizing the spherical harmonics defined on the 2D sphere to build the encoder-decoder mappings. Besides, LAINR incorporates Neural Ordinary Differential Equations (Neural ODEs) to model the latent dynamics as a continuous-time system. Importantly, the framework also offers reliable uncertainty estimations for the trained networks, which are then integrated into the assimilation cycle. The overall framework has been depicted in Figure \ref{fig:LAINR-pipeline}.

LAINR inherits the advantages of the LA framework, which allows for the integration of data-driven encoder-decoder pairs and latent surrogate models that work seamlessly with existing well-developed DA algorithms. No explicit formula for the physical dynamics is required since the latent surrogate model is also data-driven and trained from scratch. Therefore, the framework supports the application to a wide range of problems and provides versatility in assimilating complex system states.
Moreover, LAINR is much more flexible compared with the previous AutoEncoder-based LA framework in that SINRs approximate the underlying physical fields with continuous mappings and thus can handle unstructured data more naturally. The continuous modeling of the latent dynamics allows for irregular time steps, which provides a more suitable tool for practical real-world applications.

\subsection{Spherical implicit neural representations}\label{sec:SINR}
\subsubsection{Implicit neural representations}
Implicit Neural Representations (INRs) adopt a distinct approach to data modeling that deviates from traditional structure-based methods such as the AutoEncoder approach described earlier in Section \ref{sec:autoencoders}. Rather than capturing state features on discrete positions, INRs operate on coordinate-based information and model the states of interest (signals or fields) as continuous mappings, which provides a substantial advantage when working with continuous states that are sampled only at discrete grid points or mesh intersections.

The core idea of INR is to use a neural network $\bm I(\cdot,\bm z;\bm\varphi)$ to approximate a given system state $\bm u:\mR^d\to\mR^c$, where $\bm z$ stands for the corresponding latent representations and $\bm\varphi$ symbolizes the trainable network parameters. The objective is to minimize the discrepancy between $\bm I(\cdot,\bm z;\bm\varphi)$ and $\bm u$, measured with an appropriate measure $\mu$ defined on a predefined sampling set $S$. The network parameters $\bm\varphi$ can be interpreted as the consistent state characteristics and thus remain invariant across the dataset, which is to say, all the system states of the same dataset are recovered in the form of $\{\bm I(\cdot,\bm z_k;\bm\varphi)\}_{k=1}^K$ as mappings, and the latent representation $\bm z_k$ contains the temporal variations among different system states.

The training of INRs is straightforward. Let
\begin{equation}
	\bm x_k=\left\{\bm u_k(\bm p)\right\}_{\bm p\in S_k},\,k=1,2,\cdots,K
\end{equation}
be a sequence of snapshots we wish to encode, where each $\bm x_k$ is made up of samples for the system state $\bm u_k$ on the finite sampling set $S_k$ at the $k$-th step. As an example, one may choose the $L^2$ norm with the measure $\mu_k$ being a sum of Dirac delta functions $\mu_k=|S_k|^{-1}\sum_{\bm p\in S_k}\delta_{\bm p}$, which leads to a minimization problem for the function
\begin{equation}\label{eq:INR-minimization}
	\argmin_{\bm z_k,\bm\varphi}\sum_{k=1}^K\left\|\bm I(\cdot,\bm z;\bm\varphi)-\bm u\right\|_{L^2(\mu_k)}^2=\sum_{k=1}^K|S_k|^{-1}\sum_{\bm p\in S_k}|\bm I(\bm p,\bm z_k;\bm\varphi)-\bm u(\bm p)|^2.
\end{equation}
After the training process, any latent representation $\bm z$ can be decoded as the system state evaluated on an arbitrary sampling set $S$ as
\begin{equation}\label{eq:INR-decoder}
	\mD^{\textrm{INR}}(\bm z)=\bm x=\{\bm I(\bm p,\bm z;\bm \varphi)\}_{\bm p\in S}.
\end{equation}
To obtain the corresponding latent representation $\bm z'$ for a given possibly unseen observations $\bm x'$ on $S'$, we need to define the encoder mapping, which is much more challenging than the decoder given above. The encoding from $\bm x'=\{\bm u'(\bm p)\}_{\bm p\in S'}$ to $\bm z'$ is not explicitly defined and involves solving an optimization problem
\begin{equation}\label{eq:INR-encoder}
	\mE^{\textrm{INR}}\left(\bm x'\right)=\mE^{\textrm{INR}}\left(\left\{\bm u'(\bm p)\right\}_{\bm p\in S'}\right)=\bm z'=\argmin_{\bm z}\|\bm I(\cdot,\bm z;\bm\varphi)-\bm u'\|_{S'}.
\end{equation}
The high cost of the optimization problem during encoding is a major concern in most INR-based methodologies. However, our LAINR framework remains efficient since the optimization needs to be executed only for the background estimate as explained later. Meanwhile, the flexibility that the sampling set $S$ in the training and the deployment process may be different and unstructured is particularly suitable for data assimilation scenarios since the real observation data are often spatially and temporally unevenly distributed. Table \ref{tab:comp_latent_embedding} compares INR with classical ROMs.

\renewcommand{\arraystretch}{1.2}
\begin{table}[!h]
	\centering
	\begin{tabular}{c|c|c|c|c}
		\hline
		method & encoder $\mE$                                                          & decoder $\mD$                                                & linearity  & flexibility \\
		\hline
		POD    & $\bm x\mapsto \bm V^\trans\bm x$                                       & $\bm z\mapsto \bm V\bm z$                                    & linear     & fixed grid  \\
		AE     & $\bm x\mapsto f_{\bm\theta}(\bm x)$                                    & $\bm z\mapsto g_{\bm\varphi}(\bm z)$                         & non-linear & fixed grid  \\
		INR    & $\bm x\mapsto\argmin_{\bm z}\|\bm I(\cdot;\bm z,\bm\varphi)-\bm u\|_S$ & $\bm z\mapsto\{\bm I(\bm p;\bm z,\bm\varphi)\}_{\bm p\in S}$ & non-linear & mesh-free   \\
		\hline
	\end{tabular}
	\caption{Comparison of different methods for latent embedding.}
	\label{tab:comp_latent_embedding}
\end{table}
\renewcommand{\arraystretch}{1.0}

\subsubsection{Spherical filters}
Recall that a classical $L$-layer deep neural network $\bm f$ is defined by the following update formula
\begin{equation}
	\begin{aligned}
		\bm \gamma^{(0)} & =\bm x,                                                                             \\
		\bm \gamma^{(l)} & =\sigma\left(\bm W^{(l)}\bm \gamma^{(l-1)}+\bm b^{(l)}\right),\quad l=1,2,\cdots,L, \\
		\bm f(\bm x)     & =\bm \gamma^{(L)},
	\end{aligned}
\end{equation}
where each layer produces an output by applying an affine mapping followed by a non-linear activation function.
Inspired by the success of SIREN \cite{sitzmann2019siren} and Fourier Feature Networks \cite{tancik2020FourierFeatureNetworks} which make use of the trigonometric functions as the activation function and the position embeddings, Multiplicative Filter Networks (MFNs) \cite{fathony2021multiplicative} propose the multiplicative updates
\begin{equation}\label{eq:MFN-updates}
	\begin{aligned}
		\bm \gamma^{(0)} & =\bm g_0(\bm x),                                                                                 \\
		\bm \gamma^{(l)} & =\left(\bm W^{(l)}\bm \gamma^{(l-1)}+\bm b^{(l)}\right)\odot\bm g_l(\bm x),\quad l=1,2,\cdots,L, \\
		\bm f(\bm x)     & =\bm W^{(L)}\bm \gamma^{(L)}+\bm b^{(L)}
	\end{aligned}
\end{equation}
to represent the network as linear combinations of basis functions. Here, the symbol $\odot$ stands for the element-wise multiplication. For instance, by implementing the filters $\bm g_l(\bm x)$ as sinusoidal functions $\bm g_l(\bm x)=\sin(\bm\omega_l\bm x+\bm\phi_l)$, which is referred to as the ``Fourier Network'', the network output $\bm f(\bm x)$ can always be expanded as a linear combination of sinusoidal bases.
\begin{Thm}
	Each coordinate of the output of \eqref{eq:MFN-updates} is given by a finite linear combination of sinusoidal bases
	\begin{equation}
		f_j(\bm x)=\sum_{t=1}^T\bar\alpha_t^{(j)}\sin\left(\bar{\bm\omega}_t^{(j)}\bm x+\bar{\bm\phi}_t^{(j)}\right)+\bar\beta^{(j)}
	\end{equation}
	for some coefficients $\bar\alpha_t^{(j)}$, $\bar{\bm\omega}_t^{(j)}$, $\bar{\bm\phi}_t^{(j)}$ and $\bar\beta^{(j)}$ only dependent on the network parameters. \cite{fathony2021multiplicative}
\end{Thm}
Nevertheless, no convergence guarantee exists for the MFN structure with sinusoidal filters since no explicit constraint is imposed on the frequencies $\bar{\bm\omega}_t^{(j)}$ and thus the learned sinusoidal functions are not necessarily orthogonal. Meanwhile, MFN does not take advantage of the intrinsic geometry of the input space and always assumes that the input data can be extended to the whole Euclidean space.

\begin{figure}
	\centering
	\includegraphics[width=.6\textwidth]{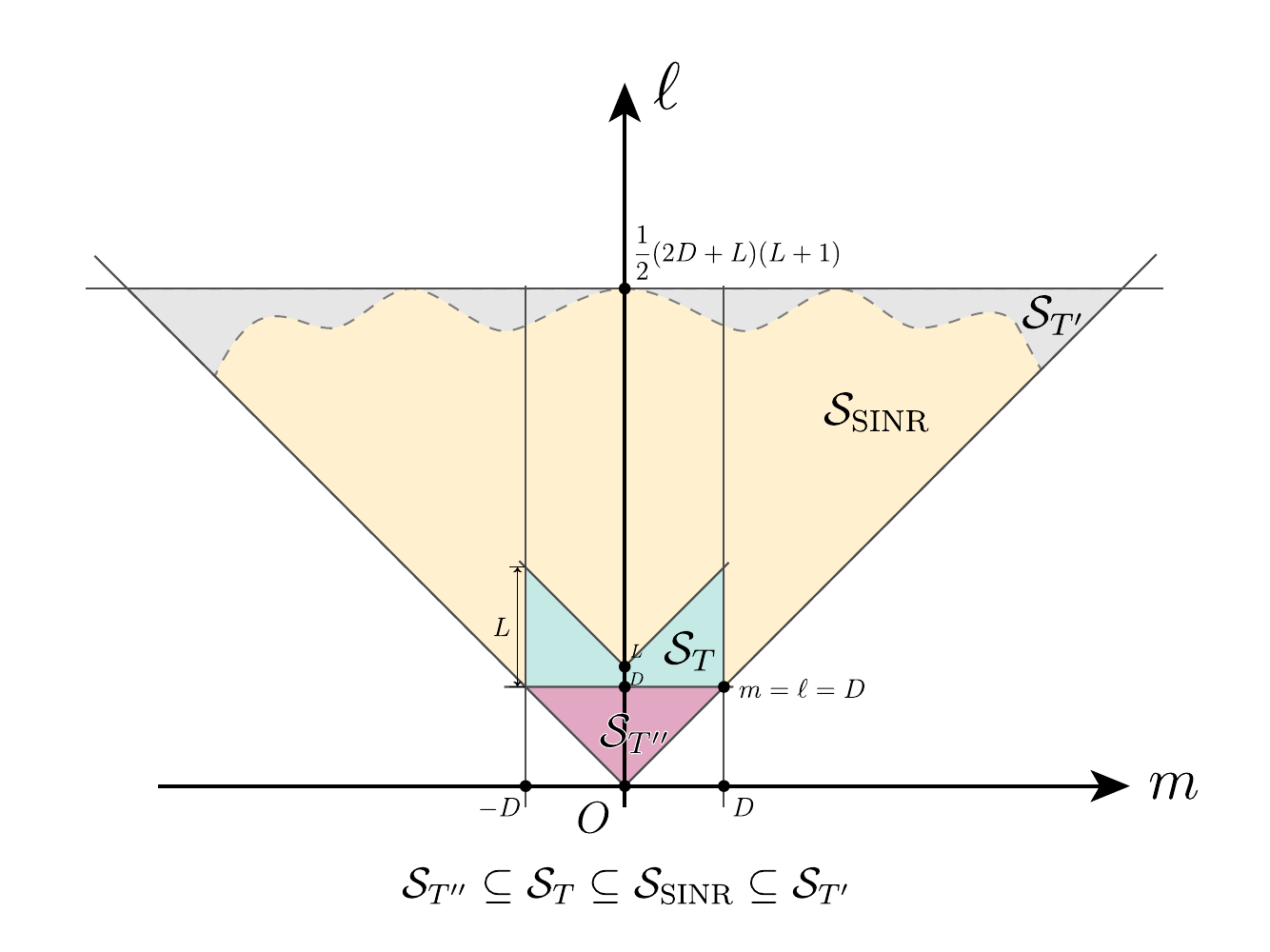}
	\caption{Illustration of the inclusion relationships  projected into the spectral modes $(m,\ell)$. Note that the range for $\mathcal{S}_{\mathrm{SINR}}$ is only for reference.}
	\label{fig:inclusion}
\end{figure}

Consequently, to deal with the input data lying on the 2D sphere, which is typically encountered in meteorological applications, we propose the spherical filter as follows
\begin{Def}[spherical filters]
	Let $\bm p=(\theta,\phi)\in\mathbb{S}^2$ be a point of the 2D sphere. A spherical filter $\bm g:\mathbb{S}^2\to\mR^h$ of degree $D$, shift $s$ and hidden dimension $h$ is defined as
	\begin{equation}\label{eq:spherical-filters}
		\bm g(\bm p)=\bm\Xi\left(Y^{-D}_{D+s}(\bm p),Y^{-(D-1)}_{D-1+s}(\bm p),\cdots, Y^0_{0+s}(\bm p),\cdots,Y^{D-1}_{D-1+s}(\bm p),Y^D_{D+s}(\bm p)\right)^\trans,\quad\bm p\in\mathbb{S}^2,
	\end{equation}
	where both $D$ and $s$ are non-negative integers and $\bm\Xi\in\mR^{h\times(2D+1)}$ is a coefficient matrix with real entries. The function $Y_\ell^m$ is the real spherical harmonics derived from the complex version
	\begin{equation}\label{eq:complex-spherical-harmonics}
		Y_{\ell,m}(\theta,\varphi)=\sqrt{\frac{2\ell+1}{4\pi}\frac{(\ell-m)!}{(\ell+m)!}}P_\ell^m(\cos\theta)\me^{\mi m\varphi}
	\end{equation}
	via
	\begin{equation}\label{eq:complex-to-real-spherical-harmonics}
		Y_\ell^m=\begin{cases}
			Y_{\ell,0}                                       & m=0, \\
			\frac{\mi}{\sqrt2}(Y_{\ell,m}-(-1)^mY_{\ell,-m}) & m<0, \\
			\frac{1}{\sqrt2}(Y_{\ell,-m}+(-1)^mY_{\ell,m})   & m>0.
		\end{cases}
	\end{equation}
	$P_\ell^m$ is the associated Legendre polynomial with the convention
	\begin{equation}\label{eq:associated-legendre-polynomial-convention}
		P_{\ell}^{-m}=(-1)^m\frac{(\ell-m)!}{(\ell+m)!}P_{\ell}^{m},\,m>0,
	\end{equation}
	and $\theta$ and $\phi$ are the polar and azimuthal angles of the spherical coordinate, respectively.
\end{Def}
\begin{Rem}
	The spherical harmonics are orthogonal basis functions on $\mathbb{S}^2$, making them a natural choice for the basis filter functions. We anticipate that this orthogonality, which has also been proved later \textcolor{modified}{in Theorem \ref{thm:convergence-SINR}}, will aid in establishing the convergence of our SINR structure. It's worth noting that in the MFN structure \eqref{eq:MFN-updates} with sinusoidal filters, both the frequencies and phase shifts of the sinusoidal functions may vary. However, we opt to fix the spherical harmonics and only train the coefficient matrix $\bm\Xi$ in \eqref{eq:spherical-filters} in our proposed spherical filters. The main reason is that prior works typically handle data lying on a certain Euclidean space, where sinusoidal functions can be easily generalized to having arbitrary real frequencies and phase shifts, especially when boundary behaviors are of little concern. Conversely, for physical states defined on the sphere $\mathbb{S}^2$, an intrinsic periodical boundary condition is introduced by the spherical geometry. Under such circumstances, the collection of spherical harmonics $\{Y_{\ell,m}\}_{\ell,m}$ can hardly be indexed by real numbers for training, as the integer parameters $\ell,m$ originate from the Legendre polynomial degrees.
\end{Rem}
Next, we focus on deriving certain convergence results for our SINR model. Fortunately, similar to the multiplication formula of trigonometric functions, any product of two spherical harmonics is also a finite sum of spherical harmonics.
\begin{Lem}\label{lem:multiplication-formula-spherical-harmonics}
	For any $|m_1|\le\ell_1$ and $|m_2|\le\ell_2$, we have
	\begin{equation}
		Y_{\ell_1}^{m_1}Y_{\ell_2}^{m_2}=\sum_{\ell=0}^{\ell_1+\ell_2}\sum_{|m|\le\ell} C_{\ell_1,\ell_2,\ell}^{m_1,m_2,m}Y_\ell^m,
	\end{equation}
	where the real coefficient $C_{\ell_1,\ell_2,\ell}^{m_1,m_2,m}$ only depends on $\ell_1,\ell_2,\ell,m_1,m_2$ and $m$.
\end{Lem}
\begin{proof}
	By the definition of the real spherical harmonics \eqref{eq:complex-to-real-spherical-harmonics}, we have for $m>0$
	\begin{equation}
		\begin{aligned}
			Y_\ell^m      & =Y_{\ell,0},                                                                              \\
			Y_{\ell}^m    & =\sqrt2(-1)^m\sqrt{\frac{2\ell+1}{4\pi}\frac{(\ell-m)!}{(\ell+m)!}}P_\ell^m\cos m\varphi, \\
			Y_{\ell}^{-m} & =\sqrt2(-1)^m\sqrt{\frac{2\ell+1}{4\pi}\frac{(\ell-m)!}{(\ell+m)!}}P_\ell^m\sin m\varphi.
		\end{aligned}
	\end{equation}
	The collection $\{Y_\ell^m\}_{|m|\le\ell}$ forms a basis of the space of real spherical harmonics of degree $\ell$ by \cite{atkinson2012spherical}. Therefore, the multiplication $Y_{\ell_1}^{m_1}Y_{\ell_2}^{m_2}$ is a polynomial of degree $(\ell_1+\ell_2)$, which is a finite linear combination of real spherical harmonics of degree no more than $(\ell_1+\ell_2)$.
\end{proof}
Different from the original MFN architecture, we add a bypass connection from each layer to the output layer in order to retain the spherical harmonics terms for lower degree $\ell$. Consequently, we propose a new variant of INRs specially designed for the 2D sphere, which is referred to as the Spherical Implicit Neural Representations (SINRs), with the detailed update formulas given below.
\begin{Def}[Spherical Implicit Neural Representations (SINRs)]
	A SINR $\bm f:\mathbb{S}^2\to\mR^c$ with $L$ layers is given by
	\begin{equation}\label{eq:SINR-updates}
		\begin{aligned}
			\bm \gamma^{(0)} & =\bm g_0(\bm p),\qquad \bm p\in\mathbb{S}^2,                                                     \\
			\bm \gamma^{(l)} & =\left(\bm W^{(l)}\bm \gamma^{(l-1)}+\bm b^{(l)}\right)\odot\bm g_l(\bm p),\quad l=1,2,\cdots,L, \\
			\bm f(\bm p)     & =\sum_{l=0}^L\left(\tilde{\bm W}^{(l)}\bm \gamma^{(l)}+\tilde{\bm b}^{(l)}\right),
		\end{aligned}
	\end{equation}
	where for each layer $\bm g_l(\bm p)$ is a spherical filter of degree $D$, shift $l$ and hidden dimension $h$ following the definition of \eqref{eq:spherical-filters}.
\end{Def}
Note that for each layer, the weights and biases $\bm W^{(l)}$, $\bm b^{(l)}$, $\tilde{\bm W}^{(l)}$ and $\tilde{\bm b}^{(l)}$, along with the coefficient matrix of the spherical filter $\bm g_l(\bm p)$, contribute to the trainable network parameters, while the number of layers $L$, the degree $D$ and the hidden dimension $h$ are hyperparameters. Similar to the Fourier Network introduced by MFN, the output of SINR can also be expanded as a linear combination of spherical harmonics.
\begin{Prop}[expansion of SINRs]\label{prop:SINR-expansion}
	Each coordinate of the output of \eqref{eq:SINR-updates} is given by a finite linear combination of real spherical harmonics. Formally, we have
	\begin{equation}
		f_j(\bm p)=\sum_{\ell=0}^{(2D+L)(L+1)/2}\sum_{|m|\le\ell}\tilde\alpha_{\ell,m}^{(j)}Y^m_\ell(\bm p)+\tilde\beta^{(j)}
	\end{equation}
	for some coefficients $\tilde\alpha_{\ell,m}^{(j)}$ and $\tilde\beta^{(j)}$ only dependent on the network parameters.
\end{Prop}
\begin{proof}
	It suffices to show that for each layer $\bm \gamma^{(l)}$, all of its coordinates can be expressed in a similar way
	\begin{equation}\label{eq:eq-to-induce}
		\gamma_j^{(l)}=\sum_{\ell=0}^{(2D+l)(l+1)/2}\sum_{|m|\le\ell}\alpha_{\ell,m}^{(j,l)}Y^m_\ell(\bm p)+\beta^{(j,l)}.
	\end{equation}
	We prove it by induction. For $l=0$, the property is trivial by the definition of $\bm g_0(\bm p)$. Suppose that \eqref{eq:eq-to-induce} holds for $l$, it immediately follows by the update
	\begin{equation}
		\bm \gamma^{(l+1)} = \left(\bm W^{(l+1)}\bm \gamma^{(l)}+\bm b^{(l+1)}\right)\odot\bm g_l(\bm p)
	\end{equation}
	that all of its coordinates can be expressed by a linear combination of products of spherical harmonics $Y_{\ell_1}^{m_1}Y_{\ell_2}^{m_2}$ where
	\begin{equation}
		|m_1|\le\ell_1\le\frac12(2D+l)(l+1),\quad|m_2|\le\ell_2\le D+l.
	\end{equation}
	By the multiplication formula of spherical harmonics (Lemma \ref{lem:multiplication-formula-spherical-harmonics}), the products can be expressed by linear combinations of spherical harmonics $Y_\ell^m$ with
	\begin{equation}
		\ell\le \frac12(2D+l)(l+1)+D+l=\frac12(2D+l+1)(l+2),
	\end{equation}
	which implies that the property holds for $l+1$.
\end{proof}
Conversely, the linear subspace spanned by the spherical harmonics can be represented by a SINR with the following proposition.
\begin{Prop}[representability of SINRs]\label{prop:SINR-representability}
	Any linear combination of spherical harmonics $Y_\ell^m$ with $|m|\le D$, $0\le\ell-|m|\le L$ can be represented accurately by a SINR \eqref{eq:SINR-updates} with $L$ layers and degree $D$ \eqref{eq:spherical-filters} as long as the hidden dimension $h\ge 2D+1$.
\end{Prop}
\begin{proof}
	By setting $\bm W^{(l)}=\mathbf{0}$, $\tilde{\bm b}^{(l)}=\mathbf{0}$ and $\bm b^{(l)}=\mathbf{1}$ for each layer, the output of \eqref{eq:SINR-updates} is given by
	\begin{equation}
		\begin{aligned}
			\bm f(\bm p) & =\sum_{l=0}^L\tilde{\bm W}^{(l)}\bm \gamma^{(l)}=\sum_{l=0}^L\tilde{\bm W}^{(l)}\bm g_l(\bm p)                                                                                         \\
			             & =\sum_{l=0}^L\tilde{\bm W}^{(l)}\bm\Xi^{(l)}\left(Y^{-D}_{D+l}(\bm p),Y^{-(D-1)}_{D-1+l}(\bm p),\cdots, Y^0_{0+l}(\bm p),\cdots,Y^{D-1}_{D-1+l}(\bm p),Y^D_{D+l}(\bm p)\right)^\trans.
		\end{aligned}
	\end{equation}
	For any linear combination of spherical harmonics $Y_\ell^m$ with $|m|\le D$, $0\le\ell-|m|\le L$ written as
	\begin{equation}
		\sum_{m=-D}^D\sum_{\ell=|m|}^{|m|+L}a_{\ell,m}Y_\ell^m(\bm p)=\sum_{l=0}^L\sum_{m=-D}^Da_{|m|+l,m}Y_{|m|+l}^m(\bm p),
	\end{equation}
	clearly there exist infinitely pairs of $(\tilde{\bm W}^{(l)},\bm\Xi^{(l)})$ such that
	\begin{equation}
		\tilde{\bm W}^{(l)}\bm\Xi^{(l)}=
		\diag\{a_{D+l,-D}, a_{D+l,-D+1}, \cdots, a_{D+l,0}, \cdots, a_{D+l,D-1}, a_{D+l,D}\}
	\end{equation}
	if $h\ge2D+1$, which completes our proof.
\end{proof}
\begin{Rem}
	In fact, the linear combinations mentioned in Proposition \ref{prop:SINR-representability} only make up a very small proportion of the set of all possible SINRs. Readers may refer to \ref{sec:SINR-representability-revisited} for another more general class of functions that can be represented by SINRs. We would like to emphasize that one of the advantages of our SINR model is the possibility of the explorations of larger degree $\ell$ with small degrees by using the graded multiplication trick, and such a feature is beneficial for a data-driven model to learn the representation of the high-frequency components of the input data.
\end{Rem}
\begin{Corl}\label{corl:inclusion-relationship}
	Denote by $\mathcal{S}_T$ the set of all linear combinations of spherical harmonics $Y_\ell^m$ with $(m,\ell)\in T$. Let
	\begin{equation}
		\begin{aligned}
			T  & =\left\{(m,\ell)\mid|m|\le D,\,0\le\ell-|m|\le L\right\},    \\
			T' & =\left\{(m,\ell)\mid|m|\le\ell\le\frac12(2D+L)(L+1)\right\}.
		\end{aligned}
	\end{equation}
	Then the set of all possible SINRs with $L$ layers, hidden dimensions $h\ge 2D+1$ and degree $D$ \eqref{eq:spherical-filters} denoted by $\mathcal{S}_{\mathrm{SINR}}$ satisfies
	\begin{equation}
		\mathcal{S}_T\subseteq\mathcal{S}_{\mathrm{SINR}}\subseteq\mathcal{S}_{T'}.
	\end{equation}
\end{Corl}
\begin{proof}
	The conclusion follows immediately from Proposition \ref{prop:SINR-expansion} and Proposition \ref{prop:SINR-representability}. Readers may refer to Figure \ref{fig:inclusion} for visualization of the inclusion relationships.
\end{proof}
By combining the expansion of the SINR architecture and its representability, we can conclude that the SINR architecture acts as a universal approximator on the 2D-sphere with the aid of the linear subspace spanned by the spherical harmonics.
\begin{Lem}[convergence of Fourier-Laplacian series]\label{lem:convergence-Fourier-Laplacian-series}
	Let $\mathcal{P}_d$ be the projection operator onto the space spanned by the spherical harmonics $Y_\ell^m$ with $|m|\le\ell=d$, then the Fourier-Laplacian series of $f\in L^2(\mathbb{S}^2)$
	\begin{equation}
		\sum_{d=0}^N\mathcal{P}_df\to f\textrm{ in }L^2(\mathbb{S}^2)\textrm{ as }N\to+\infty,
	\end{equation}
	and the convergence is uniform if $f$ is smooth enough. \cite{atkinson2012spherical}
\end{Lem}
\begin{Thm}[convergence of SINRs]\label{thm:convergence-SINR}
	For any $f\in L^2(\mathbb{S}^2)$,
	\begin{equation}
		\lim_{D,L\to\infty}\inf_{f'\in\mathcal{S}_{\mathrm{SINR}}}\|f'-f\|_{L^2(\mathbb{S}^2)}=0.
	\end{equation}
	Furthermore, if $f\in C^\infty(\mathbb{S}^2)$,
	\begin{equation}
		\lim_{D,L\to\infty}\inf_{f'\in\mathcal{S}_{\mathrm{SINR}}}\|f'-f\|_{C(\mathbb{S}^2)}=0.
	\end{equation}
\end{Thm}
\begin{proof}
	The proof is a direct consequence of the inclusion $\mathcal{S}_{T''}\subseteq\mathcal{S}_T\subseteq\mathcal{S}_{\mathrm{SINR}}$, where
	\begin{equation}
		T''=\{(m,\ell)\mid|m|\le\ell\le\min(D,L)\}
	\end{equation}
	by Corollary \ref{corl:inclusion-relationship} and Lemma \ref{lem:convergence-Fourier-Laplacian-series}.
\end{proof}
\begin{Rem}
	Qualitatively speaking, the rate of convergence is tied to the smoothness of the target function $f$. As the degree $D$ and the number of layers $L$ increase, the smoother the function is, the faster $\mathcal{S}_{T''}$ converges as well as $\mathcal{S}_{\mathrm{SINR}}$. Readers may refer to \cite{atkinson2012spherical} for detailed discussions. Generally, uniform convergence is expected since most natural physical fields exhibit smooth behaviors.
\end{Rem}
\subsection{Modulation adjustment}
\begin{figure}
	\centering
	\includegraphics[width=.8\textwidth]{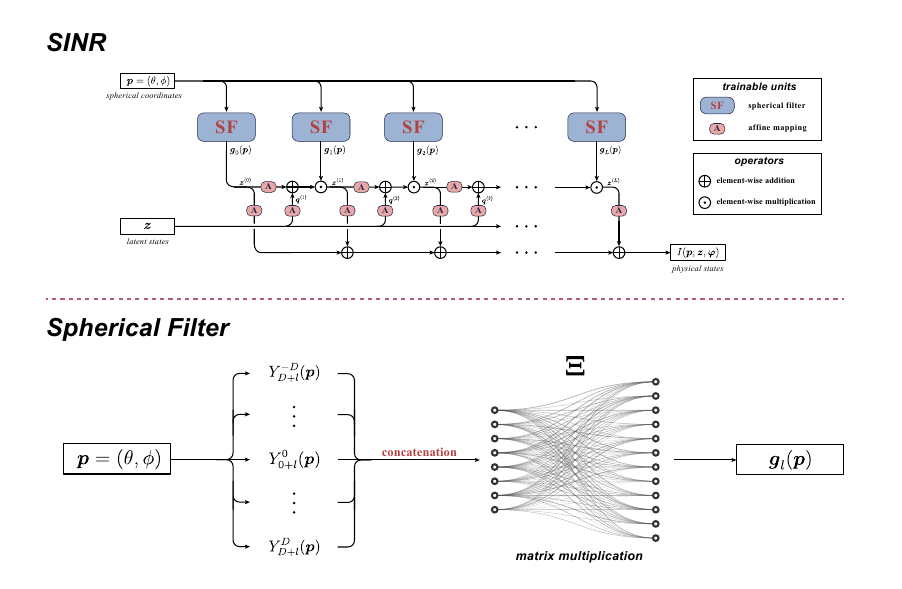}
	\caption{Visualization of the SINR architecture with modulation adjustments and the spherical filters.}
	\label{fig:SINR}
\end{figure}

Directly treating the network parameters $\bm W^{(l)}$, $\bm b^{(l)}$, $\tilde{\bm W}^{(l)}$ and $\tilde{\bm b}^{(l)}$ together with the scalar matrix of the spherical filter for each layer of our SINR architecture \eqref{eq:SINR-updates} as the latent representation $\bm z$ can hardly achieve our purpose of efficient reduction of dimensions. Therefore, we propose to use the modulation adjustment, which is also a common technique of many existing INR networks. The idea is to uses a shared base network across different snapshots to model universal physical structure, with modulations modeling the variation specific to individual time steps \cite{perez2018film,Mehta2021}. Experiments have indicated that a simple modulation shift \cite{functa22} for each layer is sufficient for our tasks. The modulation adjustment for the SINR architecture \eqref{eq:SINR-updates} is
\begin{equation}\label{eq:SINR-updates-modulation}
	\begin{aligned}
		\bm \gamma^{(0)}              & =\bm g_0(\bm p),\qquad \bm p\in\mathbb{S}^2,                                                                        \\
		\bm \gamma^{(l)}              & =\left(\bm W^{(l)}\bm \gamma^{(l-1)}+\bm b^{(l)}+\bm q^{(l)}(\bm z)\right)\odot\bm g_l(\bm p),\quad l=1,2,\cdots,L, \\
		\bm I(\bm p;\bm z,\bm\varphi) & =\sum_{l=0}^L\left(\tilde{\bm W}^{(l)}\bm \gamma^{(l)}+\tilde{\bm b}^{(l)}\right),
	\end{aligned}
\end{equation}
where $\bm q^{(l)}(\bm z)$ stands for a trainable affine mapping of the latent representation $\bm z$ for the $l$-th layer, and the parameters are not shared across different layers. The original weights and biases of the SINR structure as well as the trainable parameters of the modulation adjustment $\bm q^{(l)}$ together make up the optimization variable $\bm\varphi$ in \eqref{eq:INR-minimization}. Note that only the latent representation $\bm z$ varies across different snapshots, while the network parameters $\bm\varphi$ remain consistent. Therefore, the dimension of the latent space is reduced to the size of $\bm z$. Figure \ref{fig:SINR} visualizes the detailed structure of a SINR with modulation adjustments.
\subsection{Continuous-time latent surrogate model}\label{sec:LAINR-latent-surrogate-model}
To address the challenges posed by temporally uneven distributions of observation data, we model the latent dynamics as a continuous-time dynamical system
\begin{equation}
	\frac{\md \bm z}{\md t}=\bm F(\bm z)\approx\bm f_{\bm\psi}(\bm z)
\end{equation}
with the set of trainable parameters denoted by $\bm\psi$, and we have chosen the Neural ODE structure \cite{chen2018NeuralODE,chen2021eventfn}
\begin{equation}
	\bm z_{k+1}=\bm z_k+\int_{t_k}^{t_{k+1}}\bm f_{\bm\psi}(\bm z)\md t
\end{equation}
for the implementation. It's worth noting that such structures are suitable for handling time-series data, as they offer a data-efficient way to parameterize a trajectory over time and naturally deal with arbitrary time steps. These advantages make Neural ODEs a fitting choice for modeling the latent dynamics in our LAINR framework in that the time steps are usually irregular for realistic assimilation processes.

One may have noticed that same as Section \ref{sec:evolution}, we only consider learning the temporal latent dynamics that come from the original physical dynamics. Other constraints are set aside, such as boundary conditions, conservation laws and divergence-free conditions if they exist. One of the reasons is that we expect these additional constraints to have been implicitly encoded within the learned lower-dimensional submanifolds, which is encouraged by the optimization process requiring the encoder-decoder mapping to obtain high reconstruction accuracy at low cost. The other reason is that the trained latent surrogate model will be inserted into assimilation cycles afterward, and the defects of the predicted latent states may be corrected by the observation data to some extent.

\subsection{Modeling of uncertainties}\label{sec:modeling-uncertainties}
There are in general two types of uncertainty we need to model in the assimilation process. One is the prediction uncertainties of the surrogate model responsible for latent dynamics, which is inevitable due to the very nature of network training. The other is the uncertainty of the latent background estimate, which characterizes the latent uncertainty transformed from the physical background by the encoder mapping. We assume that the encoder mapping is perfect, which means that the optimization problem \eqref{eq:INR-encoder} is solved with a negligible margin of error. Experimental results have validated the practical effectiveness of such simplification.
\subsubsection{Uncertainty estimator for the latent surrogate model}\label{sec:uncertainty-latent-surrogate-model}
Upon successful training of the latent surrogate model, let $\hat{\bm z}_{k+1}=\mG(\bm z_k)$ be the predicted latent state for the $(k+1)$-th time step based on the $k$-th latent state $\bm z_k$. We model the prediction uncertainty as a Gaussian distribution. To put it formally,
\begin{equation}
	\mG(\bm z_k)=\hat{\bm z}_{k+1}=\bm z_{k+1}+\bm\varepsilon_{k+1}^L,\quad\bm\varepsilon_{k+1}^L\sim\mathcal{N}\left(0,\bm\Sigma^L\right),
\end{equation}
where $\bm\Sigma^L$ is the covariance matrix of the uncertainty $\bm\varepsilon_{k+1}^L$ of the latent surrogate model $\mG$, and we assume it to be time-independent.

We have to point out, however, that our choice of using Gaussian models is far from optimality. We only aim to argue for the necessity of such a mechanism in our LAINR framework, which is capable of estimating the uncertainty of the learned latent surrogate model and thus benefiting the assimilation process. The Gaussian error modeling for the surrogate model depends on the model's adequate representability for the latent dynamics, and we assume that the model error only results from unknown random forcing or random variation of parameters. Nonetheless, such an assumption does not necessarily hold in general. Literature such as \cite{Farchi2021} has employed specialized and advanced models to characterize the possible systematical biases of the surrogate model, which is not the main focus of our current work.

To train the uncertainty estimator, our LAINR framework assumes that the $\bm\Sigma^L$ is diagonal, which seems to be quite arbitrary, but experimental results have surprisingly shown its effectiveness (Section \ref{sec:exp-uncertainty-estimation}). We parameterize $\left(\bm\Sigma^L\right)^{1/2}=\exp\bm D$, and $\exp\bm D$ stands for the element-wise exponential of a diagonal matrix $\bm D$ to ensure its positive-definiteness. For convenience, we refer to the matrix $\bm D$ as our uncertainty estimator for the latent surrogate model hereafter. The matrix $\bm D$ is initialized with zeros and then optimized during training, where the target loss function is derived as follows.
With the independence assumption, the likelihood of the uncertainties $\{\bm\xi_k\}_k$ across the whole dataset reads
\begin{equation}
	\mathbb{P}\left(\{\bm\xi_k\}_k\right)=\prod_k(2\pi)^{-m/2}\det\left(\bm\Sigma^L\right)^{-1/2}\exp\left(-\frac12\left(\mG(\bm z_k)-\bm z_{k+1}\right)^\trans\left(\bm\Sigma^L\right)^{-1}\left(\mG(\bm z_k)-\bm z_{k+1}\right)\right).
\end{equation}
Consequently, we train the uncertainty estimator $\bm D$ by minimizing the negative log-likelihood
\begin{equation}
	\begin{aligned}
		\mathcal{L}_\textrm{MLE} & =\sum_k\frac12\log\det\left(\bm\Sigma^L\right)+\frac12\left(\mG(\bm z_k)-\bm z_{k+1}\right)^\trans\left(\bm\Sigma^L\right)^{-1}\left(\mG(\bm z_k)-\bm z_{k+1}\right) \\
		                         & =\sum_k\log\det(\exp\bm D)+\frac12\left(\mG(\bm z_k)-\bm z_{k+1}\right)^\trans(\exp\bm D)^{-1}(\exp\bm D)^{-1}\left(\mG(\bm z_k)-\bm z_{k+1}\right)                  \\
		                         & =\sum_k\left(\sum_j\bm D_{jj}+\frac12\left\|(\exp\bm D)^{-1}\left(\mG(\bm z_k)-\bm z_{k+1}\right)\right\|^2\right)
	\end{aligned}
\end{equation}
up to a constant by the method of maximum likelihood estimation (MLE), where the outer summation runs through all the latent representations $\bm z_k$, and the inner summation $\sum_j\bm D_{jj}$ represents the trace of $\bm D$.
\subsubsection{Uncertainty of the latent background estimate}\label{sec:uncertainty-latent-background-estimate}
Consider a physical background estimate
\begin{equation}
	\bm x_0^b=\bm x_0+\bm\varepsilon_x^b,\quad\bm\varepsilon_x^b\sim\mathcal{N}(0,\bm\Sigma_x^b)
\end{equation}
given as a Gaussian perturbation of the true physical state $\bm x_0$.
The encoder mapping $\mE$ transforms it into a latent background estimate
\begin{equation}
	\bm z_0^b=\mE\left(\bm x_0^b\right)=\mE\left(\bm x_0+\bm\varepsilon_x^b\right)=\mE\left(\bm x_0\right)+\bm J_\mE(\bm x_0)\bm\varepsilon_x^b+o\left(\bm\varepsilon_x^b\right)=\bm z_0+\bm J_\mE(\bm x_0)\bm\varepsilon_x^b+o\left(\bm\varepsilon_x^b\right),
\end{equation}
where $\bm J_\mE$ is the Jacobian matrix of $\mE$. By omitting the high-order term, we make the following approximation
\begin{equation}
	\bm z_0^b\approx\bm z_0+\bm J_\mE(\bm x_0)\bm\varepsilon_x^b=\bm z_0+\bm\varepsilon_z^b,\quad\bm\varepsilon_z^b\sim\mathcal{N}\left(0,\bm J_\mE(\bm x_0)\bm\Sigma_z^b\bm J_\mE(\bm x_0)^\trans\right).
\end{equation}
Since the encoder mapping $\mE$ is associated with an optimization process, no explicit formula for the Jacobian matrix $\bm J_\mE$ is available. Therefore, we choose to make use of the decoder mapping $\mD$ in that the chain rule gives
\begin{equation}
	\bm J_\mE(\bm x_0)\bm J_\mD(\bm z_0)=\bm J_\mE(\mD(\bm z_0))\bm J_\mD(\bm z_0)=\bm J_{\mE\circ\mD}(\bm z_0)=\bm I_m.
\end{equation}
First, for any physical background estimate $\bm x_0^b$, LAINR calculates the latent background estimate $\bm z_0^b=\mE\left(\bm x_0^b\right)$. It follows that the Jacobian $\bm J_\mD(\bm z_0)$ is approximated by $\bm J_\mD\left(\bm z_0^b\right)$ with the aid of automatical differentiation on the decoder mapping $\mD$. The Moore-Penrose inverse of $\bm J_\mD\left(\bm z_0^b\right)$ is used to approximate $\bm J_\mE(\bm x_0)$ afterwards in that they are both the left inverse of $\bm J_\mD\left(\bm z_0^b\right)$, which is then employed to construct the covariance matrix $\bm J_\mE(\bm x_0)\bm\Sigma_z^b\bm J_\mE(\bm x_0)^\trans$ for $\bm\varepsilon_z^b$.

\textcolor{black}{We need to address that the uncertainty of the latent background estimate is only used for start-up, and all the subsequent assimilation cycles are conducted within the latent space. The uncertainty $\bm\Sigma_x^b$ for the physical state is commonly accessible since the background state usually comes from the forecast starting from the past analysis state. 
	Besides, the uncertainty $\bm\Sigma_x^b$ does not need to be very accurate since, in the context of data assimilation, it is the consistent observations at different time steps, rather than the initial conditions that calibrate the assimilated model state to keep evolving closely to the true state.}
\subsection{The proposed LAINR framework}
LAINR shares some similarities with the LA framework described in Section \ref{sec:LA-framework}. It requires observed data $\{\bm y_k\}_k$, observation operators $\{\mH_k\}_k$ and a background estimate $\bm x_0$ to initiate the assimilation process. In contrast to the LA framework, any unstructured historical dataset encoding the physical dynamics is sufficient for LAINR to obtain a latent surrogate model for the dynamics, and LAINR additionally provides the corresponding uncertainty estimations for the trained networks. The overall procedure for the deployment of our LAINR framework is summarized as follows.
\begin{enumerate}[(i)]
	\item Train the encoder-decoder mapping implemented via the SINR architecture as detailed in Section \ref{sec:SINR} with \eqref{eq:INR-minimization}. The resulting encoder $\mE:\mX'\to\mZ$ and decoder $\mD:\mZ\to\mX'$ are given by \eqref{eq:INR-encoder}\eqref{eq:INR-decoder}. Besides, the latent surrogate model $\mG:\mZ\to\mZ$ is implemented via Neural ODE (Section \ref{sec:LAINR-latent-surrogate-model}) such that
	      \begin{equation}
		      \mD\circ\mE\approx\textrm{identity mapping on }\mX';\qquad\mD\circ\mG\circ\mE\approx\mM.
	      \end{equation}
	      In other words, the encoder $\mE$ and the decoder $\mD$ are inverse mappings such that the diagram in Figure \ref{fig:comp-DA-and-LA} (b) commutes up to a small error.
	\item Use the uncertainty estimator to evaluate the uncertainty covariance matrix $\bm\Sigma^L$ for the latent surrogate model $\mG$. The uncertainty estimator is trained by the method of maximum likelihood estimation (MLE) using the latent representation $\{\bm z_k\}_k$ of the training dataset. (Section \ref{sec:uncertainty-latent-surrogate-model})
	\item Determine the latent background estimate $\bm z_0^b=\mE\left(\bm x_0^b\right)$ and calculate its associated uncertainty covariance matrix. (Section \ref{sec:uncertainty-latent-background-estimate})
	\item Implement the assimilation process on the latent space with an existing well-developed DA algorithm, where $\mG$ serves as the latent forward propagation operator and $\mH'=\mH\circ\mD$ acts as the observation operator. The observation noise of $\mH'=\mH\circ\mD$ remains the same as that of $\mH$. The uncertainty estimations are given in the previous two steps.
	\item Retrieve the assimilated system states $\left\{\bm x_k^a\right\}_k$ from their assimilated latent representations via the decoder $\bm x_k^a=\mD\left(\bm z_k^a\right)$.
\end{enumerate}

\section{Experiments}\label{sec:experiments}
To assess the capability of our proposed LAINR framework, we have conducted multiple experiments to make comparisons with existing works based on AutoEncoders. Additionally, we have studied the flexibility of our LAINR framework when handling various inconsistent data structures as well.

First, the test cases we have adopted in this paper are introduced, consisting of an ideal shallow-water dataset and a more realistic ERA5 reanalysis dataset to explore the model applicability. After that, the model configurations including the structures and the hyper-parameters along with the metrics for training and evaluation are exhibited, which are then followed by the details of the training process.

\begin{figure}[ht]
	\centering
	\begin{subfigure}{.45\textwidth}
		\centering
		\includegraphics[width=.6\textwidth]{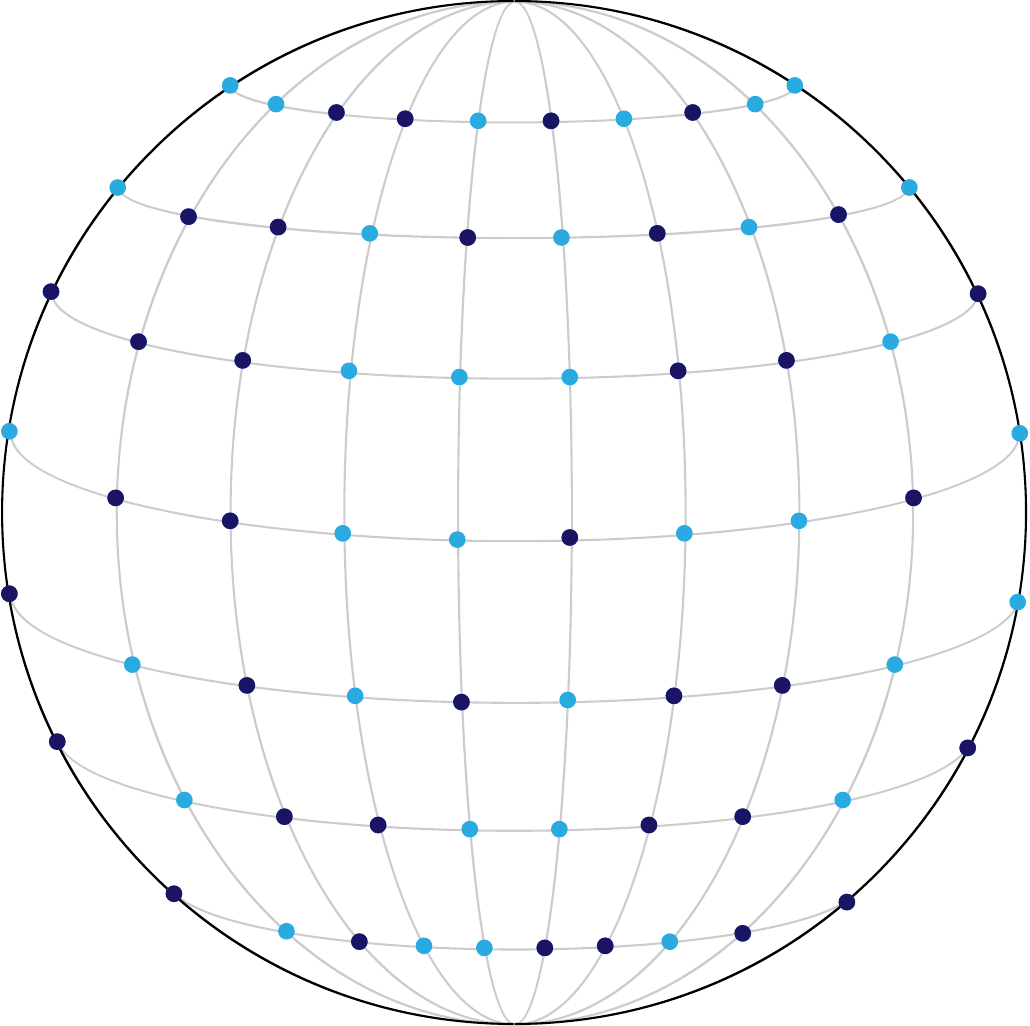}
		\caption{The grid $G$ (dark \& light blue) is used for both training and assimilation when we make comparisons with AutoEncoder-based methods. The dark blue points represent the possible observed locations $S_{\textrm{obs}}$ during assimilation.}
		\label{fig:regular-grid}
	\end{subfigure}
	\hfill
	\begin{subfigure}{.45\textwidth}
		\centering
		\includegraphics[width=.6\textwidth]{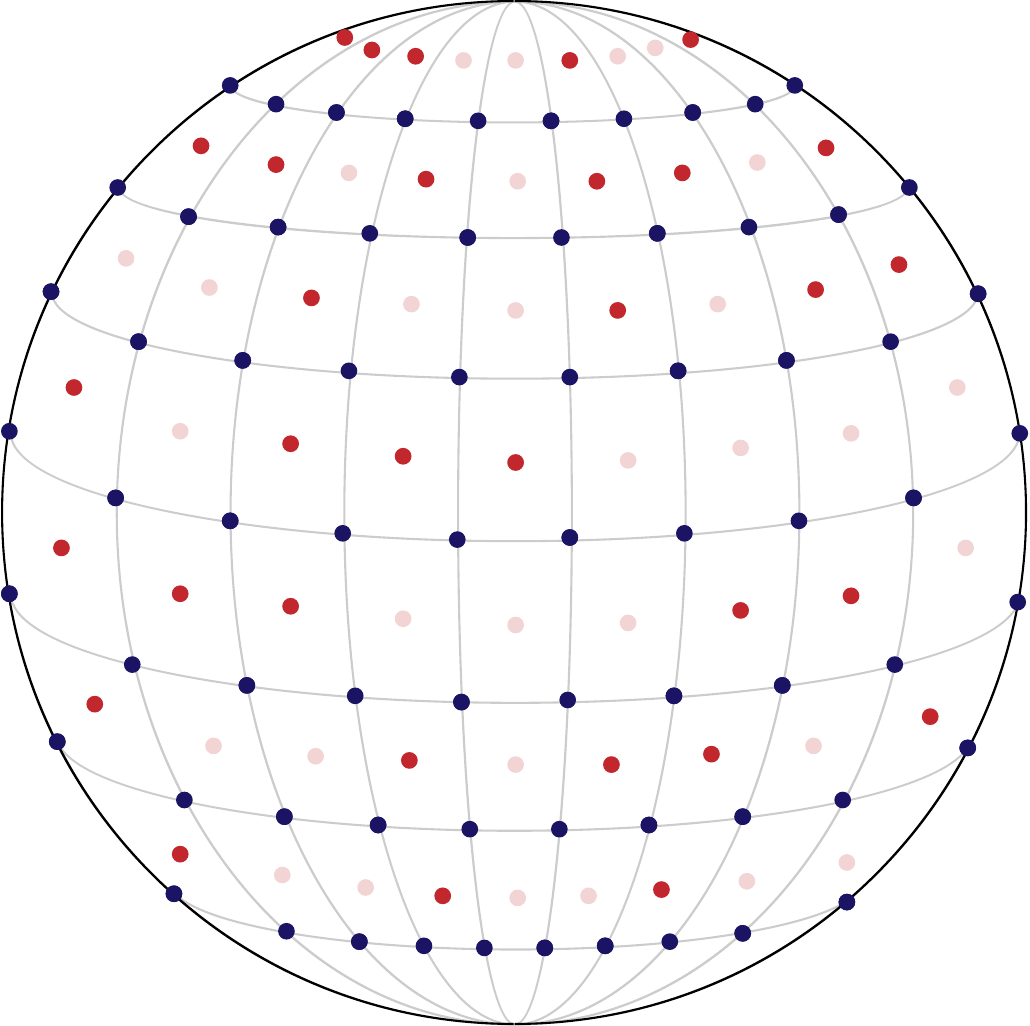}
		\caption{The grid $G$ (dark blue) is used for training and the staggered grid $G'$ (dark \& light red) is used for assimilation to study the flexibility of LAINR. The dark red points represent the possible observed locations $S_{\textrm{obs}}'$ during assimilation.}
		\label{fig:staggered-grid}
	\end{subfigure}
	\caption{Grids for training and assimilation}
	\label{fig:grid-illustration}
\end{figure}

The detailed experiments are presented in Section \ref{sec:performance-comparisons}, where we have compared our LAINR framework with the existing AutoEncoder-based methodologies. Both the training and assimilation processes are based on a regular latitude-longitude grid (Figure \ref{fig:regular-grid}). The comparisons are made in terms of the reconstruction, the prediction as well as the assimilation accuracy. All the experimental results have demonstrated the superiority of our LAINR framework over the existing AutoEncoder-based methodologies.

Furthermore, LAINR has the flexibility to work with diverse types of sampling sets with little adjustment, and we have examined the flexibility of our LAINR framework in Section \ref{sec:mesh-free-flexibility} by testing on a reconstruction task with only rare available data, and a zero-shot assimilation task where the training and testing datasets come from different grids (Figure \ref{fig:staggered-grid}). Our LAINR framework has shown its large potential in handling incomplete data and unseen sampling sets.

\subsection{Test cases and model configurations}
\subsubsection{Test cases}\label{sec:test-eqs}
We choose multi-dimensional partial differential equations (PDEs) as our test cases instead of chaotic ordinary differential equations (ODEs) such as the Lorenz-96 model adopted in the previous works \cite{Peyron2021LAwithAE,LatentspaceDA-RNN} to study the performances of our LAINR framework. PDEs are more common in nature and technology, and they typically possess much more complex physical dynamics than ODEs. For this study, both the ideal shallow-water model as well as the more realistic ERA5 reanalysis data have been established as the test cases in this paper.

The first one is the ideal shallow-water model, which is a simplified version of the primitive equations of the atmosphere. The second one is the ERA5 reanalysis datasets, which are the latest global atmospheric reanalysis datasets produced by the European Centre for Medium-Range Weather Forecasts (ECMWF). A brief description of the two test cases is given below, and readers may refer to \ref{sec:dataset_configs} for detailed dataset configurations.

It is also important to mention that similar to most of the previous DL-based assimilation frameworks such as \cite{Peyron2021LAwithAE,LatentspaceDA-RNN,ROM-DA,GeneralizedLA}, our LAINR framework and the AutoEncoder baselines do not assume any prior knowledge of the underlying dynamics. We only extract partial physical features from each PDE system to build the datasets used for model training to impose a much more difficult setting.

\paragraph{shallow-water dataset}
We consider the following spherical shallow-water equation on the 2D sphere
\begin{equation}
	\begin{aligned}
		\frac{\md\bm u}{\md t} & =-f\bm k\times\bm u-g\nabla h+\nu\Delta\bm u, \\
		\frac{\md h}{\md t}    & =-h\nabla\cdot\bm u+\nu\Delta h,
	\end{aligned}
\end{equation}
as a simplification of the primitive equations of atmospheric flow with the material derivative denoted by $\frac{\md}{\md t}$. The vorticity field $w=\nabla\times\bm u$ together with the thickness of the fluid layer $h$ becomes the system states to be assimilated. $\nu$ stands for the viscosity, and the parameters $f$, $g$, $\Omega$ are all fixed and in consistent with the real earth surface. 20 trajectories starting from different initial states have been generated and downsampled onto a $128\times64$ regular latitude-longitude grid with $\Delta t\approx1$ real hour. All the trajectories have been truncated into 240 time steps, and 10\% of them are separately used for testing purposes, which means the shapes of the training and testing datasets are $(18, 240,128,64,2)$ and $(2,240,128,64,2)$, respectively.

\paragraph{ERA5 dataset}
The ERA5 dataset \cite{ERA5} is the latest global atmospheric reanalysis dataset produced by the European Centre for Medium-Range Weather Forecasts (ECMWF), which is generated by assimilating observations from across the world into a delicate numerical weather prediction model. In our experiments, we do not make comparisons with the DA algorithms employed in the generation of the ERA5 dataset, but treat the dataset as the ground truth to provide a more realistic dynamical system than the previous ideal shallow-water dataset. We select the Z500 (geopotential at 500 hPa) and the T850 (temperature at 850 hPa) fields to construct our dataset, and we still refer to it as the ``ERA5 dataset'' for convenience. The choice of Z500 and T850 aligns with those of the early attempts \cite{Clare2021,Scher2021Ensemble,Weyn2020,Rasp2021} of DL-based weather forecasting, which in our setting, become the physical features we aim to assimilate. The spatial resolution of the Z500 and T850 fields are $128\times64$ with a 1-hour time interval. We have extracted 10 trajectories of a whole year for training, and 2 additional trajectories for testing purposes, which produce datasets of size approximately $(10,365\times24,128,64,2)$ and $(2,365\times24,128,64,2)$, respectively.
\subsubsection{Model configurations}\label{sec:model-configs}
\paragraph{AutoEncoder}
In the previous work \cite{Peyron2021LAwithAE}, the encoder-decoder mappings have been implemented with fully-connected neural networks, but such design choice raises potential scalability issues for our two-dimensional test cases, considering the prohibitively large parameter volume. Directly applying fully-connected networks could lead to a parameter volume of roughly $(2\times128\times64)^2\times d\approx2.7\times10^8\times d$ for a network of depth $d$ with 2-channel input of size $128\times64$, which is impractical, particularly as this volume grows with the scale of the real deployment.
Consequently, to make a systematical comparison with the AutoEncoder-based LA framework and circumvent the scalability concerns, we have opted to leverage convolutional neural networks (CNNs) to establish our baselines. The decision aligns with the natural suitability of CNNs for tasks that exhibit local spatial or temporal correlations.

We have chosen the Convolutional AutoEncoder (\textbf{CAE}) structure as employed in \cite{Amendola2021CAE-DA} for assimilating the room $\textrm{CO}_2$ concentration, and the \textbf{AEflow} \cite{AEflow} structure proposed in the field of turbulent flow simulation. With the latent dimension fixed, we vary the kernel size $K\in\{3,5,7\}$ and the number of hidden channels $C\in\{16,32,64\}$ for the CAE structure, and the kernel size $K\in\{3,5,7\}$ and the number of residual blocks $N_{\textrm{res}}\in\{4,8,12\}$ for the AEflow structure. To make the convolutions compatible with the boundary condition, we utilize periodic paddings along the zonal directions and replication paddings at the poles. See \ref{sec:net-structures} for detailed descriptions.


\paragraph{INR}
As introduced in Section \ref{sec:SINR}, we have chosen the SINR structure as our INR implementation due to its reliable convergence guarantee. By empirically tuning the hyperparameters, we fix the hidden dimensions $h=128$ for each state variable and $D=L=8$ for our implementation of SINR in the following experiments.
\subsubsection{Metrics}
To show the advantages of our LAINR framework, it is important to quantify the performances by either the rooted-mean-square error (RMSE) evaluated on a pre-defined quasi-uniform skipped latitude-longitude grid \cite{Weller2012SkipLatLon}, or the rooted-mean-square error (RMSE) weighted by the cosine values of latitudes. We choose to employ the latter one since it is more compatible with the $L^2$ norm on the sphere and more flexible when applied to unstructured data. The weighted RMSE is formulated as
\begin{equation}
	\mL_{\textrm{sqrt}}(\bm x,\bm x')=\sqrt{\frac{1}{\sum_{\bm p\in G}\cos\phi(\bm p)}\sum_{\bm p\in G}\left\|\bm x(\bm p)-\bm x'(\bm p)\right\|^2\cos\phi(\bm p)},
\end{equation}
where $\phi(\bm p)$ stands for the latitude of the position $\bm p$, and all the RMSEs mentioned hereafter are computed as above unless otherwise specified.
Besides, we also define an ``unrooted'' version
\begin{equation}
	\mL(\bm x,\bm x')=\frac{1}{\sum_{\bm p\in G}\cos\phi(\bm p)}\sum_{\bm p\in G}\left\|\bm x(\bm p)-\bm x'(\bm p)\right\|^2\cos\phi(\bm p),
\end{equation}
for network training, and the details can be found in the following subsections.
\subsubsection{Training process}
We divide the training process into 2 stages, namely the \textit{pre-training} stage and the \textit{fine-tuning} stage. The pre-training stage is designed to obtain a good latent representation for all of the training snapshots, and the fine-tuning stage focuses mainly on the reconstruction of the latent dynamics.
\paragraph{Pre-training}
We need to emphasize that the trainable parameters for the AutoEncoder baselines are the network parameters of both their encoders and decoders. On the contrary, no explicit encoding mapping is needed during the INR training, and the trainable parameters are made up of the decoder parameters as well as the latent representations.
The reconstruction loss is used to measure the discrepancy between the original data and the data reconstructed from the latent representations. Naturally, we define the reconstruction loss for the AutoEncoder baselines as
\begin{equation}\label{eq:ae_reconstruction_loss}
	L_{\textrm{rec}}^{\textrm{AE}}(\bm x)=\mL(\bm x,\mD_{\bm\varphi}\circ\mE_{\bm\theta}(\bm x)).
\end{equation}
However, since the latent representations are also trainable parameters for the INR models, the corresponding reconstruction loss is a little different:
\begin{equation}\label{eq:inr_reconstruction_loss}
	L_{\textrm{rec}}^{\textrm{INR}}(\bm x,\bm z)=\mL(\bm x,\mD_{\bm\varphi}(\bm z)).
\end{equation}
Consequently, for the training dataset $\{\bm x_k\}_{k=1}^K$, the pre-training stage solves
\begin{equation}\label{eq:ae_reconstruction_loss_empirical}
	\argmin_{\bm\theta,\bm\varphi}\frac1K\sum_{k=1}^KL_{\textrm{rec}}^{\textrm{AE}}(\bm x_k)=\argmin_{\bm\theta,\bm\varphi}\frac1K\sum_{k=1}^K\mL(\bm x_k,\mD_{\bm\varphi}\circ\mE_{\bm\theta}(\bm x))
\end{equation}
and
\begin{equation}\label{eq:inr_reconstruction_loss_empirical}
	\argmin_{\bm z_k,\bm\varphi}\frac1K\sum_{k=1}^KL_{\textrm{rec}}^{\textrm{INR}}(\bm x_k,\bm z_k)=\argmin_{\bm\theta,\bm\varphi}\frac1K\sum_{k=1}^K\mL(\bm x_k,\mD_{\bm\varphi}(\bm z_k))
\end{equation}
for the AutoEncoder baselines and the INR models, respectively.
\paragraph{Fine-tuning}
To train a latent surrogate model to approximate the unknown latent dynamics, we define the multi-step prediction errors for the $s$-th step as
\begin{equation}\label{eq:prediction_loss}
	L_{\textrm{pred}}^{\textrm{AE},(s)}(\bm x_k)=\left\|\mE_{\bm\theta}(\bm x_{k+s})-\mG_{\bm\psi}^s\circ\mE_{\bm\theta}(\bm x_k)\right\|^2,\,L_{\textrm{pred}}^{\textrm{INR},(s)}(\bm z_k)=\left\|\bm z_{k+s}-\mG_{\bm\psi}^s(\bm z_k)\right\|^2
\end{equation}
with the common mean-squared errors on the latent space. The choice of $s$ is a bit tricky, as a smaller $s$ will force the models to learn the next-step prediction behavior and ignore the long-term dynamics, while a larger $s$ may result in difficulties in training.
Nevertheless, in our test cases, we have found that a fixed $s$ is sufficient for the surrogate models to learn effective latent dynamics used for assimilation.
The training process of the latent surrogate models is formulated as
\begin{equation}\label{eq:ae_prediction_loss_empirical}
	\argmin_{\bm\psi}\frac1K\sum_{k=1}^KL_{\textrm{pred}}^{\textrm{AE},(s)}(\bm x_k)=\argmin_{\bm\psi}\frac1K\sum_{k=1}^K\left\|\mE_{\bm\theta}(\bm x_{k+s})-\mG_{\bm\psi}^s\circ\mE_{\bm\theta}(\bm x_k)\right\|^2
\end{equation}
and
\begin{equation}\label{eq:inr_prediction_loss_empirical}
	\argmin_{\bm\psi}\frac1K\sum_{k=1}^KL_{\textrm{pred}}^{\textrm{INR},(s)}(\bm z_k)=\argmin_{\bm\psi}\frac1K\sum_{k=1}^K\left\|\bm z_{k+s}-\mG_{\bm\psi}^s(\bm z_k)\right\|^2
\end{equation}
for the AutoEncoder baselines and the INR models, respectively. Note that in the fine-tuning stage, the optimization problems \eqref{eq:ae_reconstruction_loss_empirical}\eqref{eq:inr_reconstruction_loss_empirical} are solved as well, but with a smaller learning rate.

The detailed training procedures for the AutoEncoder models and the INR models are summarized in Algorithm \ref{alg:aeflow_rezero} and Algorithm \ref{alg:inr_neuralode}, respectively, where we demonstrate the optimization with Stochastic Gradient Descent (SGD) with a consistent learning rate $\eta$ for simplicity.

\begin{minipage}{.45\textwidth}
	\begin{algorithm}[H]
		\centering
		\caption{Training for AutoEncoders}
		\label{alg:aeflow_rezero}
		\footnotesize
		\begin{algorithmic}[1]
			\STATE {\textbf{Input:} Training data $\{\bm x_k\}_k$, epochs $T$}
			\STATE {\textbf{Output:} trained models $\mE_{\bm\theta}$, $\mD_{\bm\varphi}$, $\mG_{\bm\psi}$}
			\STATE Initialize $\mE_{\bm\theta}$, $\mD_{\bm\varphi}$ and $\mG_{\bm\psi}$.
			\vspace*{5ex}
			\FOR{$t = 1$ to $T$}
			\FOR{$\bm x_k$ in $\{\bm x_k\}_k$}
			\STATE \textit{** forward propagation **}
			\STATE Calculate the reconstruction loss $L_{\textrm{rec}}^{\textrm{AE}}(\bm x_k)$;
			\STATE Calculate the prediction loss $L_{\textrm{pred}}^{\textrm{AE},(s)}(\bm x_k)$;
			\ENDFOR
			\STATE \textit{** update model weights **}
			\STATE $\bm\theta\leftarrow\bm\theta-\eta\nabla_{\bm\theta}\sum_k\left(L_{\textrm{rec}}^{\textrm{AE}}(\bm x_k)+L_{\textrm{pred}}^{\textrm{AE},(s)}(\bm x_k)\right)$
			\STATE $\bm\varphi\leftarrow\bm\varphi-\eta\nabla_{\bm\varphi}\sum_k\left(L_{\textrm{rec}}^{\textrm{AE}}(\bm x_k)+L_{\textrm{pred}}^{\textrm{AE},(s)}(\bm x_k)\right)$
			\STATE $\bm\psi\leftarrow\bm\psi-\eta\nabla_{\bm\psi}\sum_kL_{\textrm{pred}}^{\textrm{AE},(s)}(\bm x_k)$
			\STATE Update $s$.
			\ENDFOR
		\end{algorithmic}
	\end{algorithm}
\end{minipage}
\hfill
\begin{minipage}{.45\textwidth}
	\begin{algorithm}[H]
		\caption{Training for INRs}
		\label{alg:inr_neuralode}
		\footnotesize
		\begin{algorithmic}[1]
			\STATE {\textbf{Input:} Training data $\{\bm x_k\}_k$, epochs $T$}
			\STATE {\textbf{Output:} trained models $\mD_{\bm\varphi}$, $\mG_{\bm\psi}$}
			\STATE Initialize $\mD_{\bm\varphi}$, $\mG_{\bm\psi}$ and $\{\bm z_k\}_k$.
			\vspace*{2ex}
			\FOR{$t = 1$ to $T$}
			\FOR{$\bm x_k$ in $\{\bm x_k\}_k$}
			\STATE \textit{** train the encoder-decoder **}
			\STATE Calculate the reconstruction loss $L_{\textrm{rec}}^{\textrm{INR}}(\bm x_k,\bm z_k)$;
			\STATE Calculate the prediction loss $L_{\textrm{pred}}^{\textrm{INR,(s)}}(\bm z_k)$;
			\STATE \textit{** update only the latent vectors **}
			\STATE $\bm z_k\leftarrow\bm z_k-\eta\nabla_{\bm z_k}L_{\textrm{rec}}^{\textrm{INR}}(\bm x_k,\bm z_k)$
			\ENDFOR
			\STATE \textit{** update model weights **}
			\STATE $\bm\varphi\leftarrow\bm\varphi-\eta\nabla_{\bm\varphi}\sum_kL_{\textrm{rec}}^{\textrm{INR}}(\bm x_k,\bm z_k)$
			\STATE $\bm\psi\leftarrow\bm\psi-\eta\nabla_{\bm\psi}\sum_kL_{\textrm{pred}}^{\textrm{INR},(s)}(\bm z_k)$
			\STATE Update $s$.
			\ENDFOR
			\vspace*{2ex}
			\STATE \textit{** output the encoder **}
			\STATE the encoder $\mE$ is given implicitly as follows:
			\STATE $\bm x\mapsto\bm z=\argmin_{\bm z'}\mL(\bm x_k,\mD_{\bm\varphi}(\bm z'))$.
		\end{algorithmic}
	\end{algorithm}
\end{minipage}
\subsection{Performance comparisons}\label{sec:performance-comparisons}
\subsubsection{Exploring suitable encoder-decoder structures}
Our experiments start with an exploration of suitable encoder-decoder structures for the shallow-water dataset, by making a comparative analysis of the reconstruction capabilities of the three distinct ROMs: CAE, AEflow and SINR. The assessment has been conducted independently of the surrogated models assigned to handle latent dynamics before the fine-tuning stage.


We set the latent dimension for the AutoEncoder baselines (CAE and AEflow) as 1024 to obtain minimal modifications from the original proposed structure \cite{Amendola2021CAE-DA,AEflow} and that for the INR encoders as 400, which presents a more challenging setup.
Furthermore, to perform an ablation study on the superiority of our proposed SINR architecture among other existing INR models, we additionally implement
\begin{itemize}
	\item \textbf{FourierNetCartes} \cite{yin2023dino}, a Fourier Network structure \cite{fathony2021multiplicative} fed with the 3D cartesian coordinates,
	\item \textbf{FourierNetLatlon}, a Fourier Network structure \cite{fathony2021multiplicative} fed with the 2D latitude-longitude coordinates, and
	\item \textbf{SINRNoSkip}, a SINR structure without the skip connections, where all the terms of the last summation in \eqref{eq:SINR-updates} are removed except for the last layer.
\end{itemize}

To make a fair comparison with our proposed framework, we have also conducted a grid search (\ref{sec:grid-search}) on the hyperparameters (the network depth, the convolutional kernel size, etc.) for the previous CAE and AEflow architectures. The optimal network configurations are fixed and used for all the following experiments.

\begin{figure}
	\centering
	\includegraphics[width=.5\textwidth]{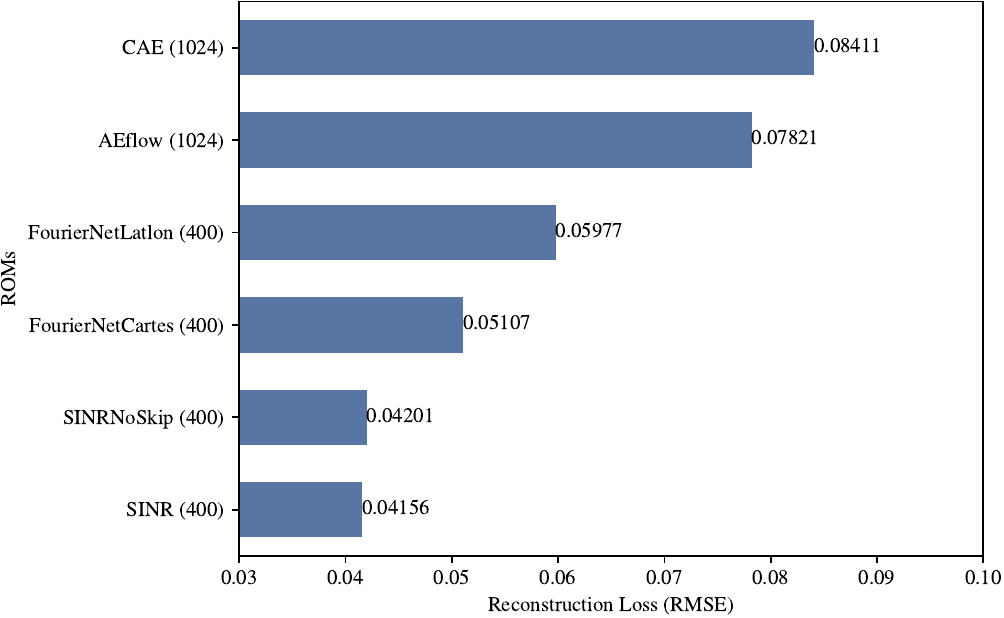}
	\caption{Comparison of the capabilities for low-dimensional reconstructions on the shallow-water dataset}
	\label{fig:comp_eff}
\end{figure}

The reconstruction errors evaluated on the testing datasets are summarized in Figure \ref{fig:comp_eff}, where the numbers of latent dimensions are appended to the method names for clarification. Even with a higher latent dimension, the AutoEncoder baselines have obtained relatively lower accuracy, indicating their inefficiency in capturing and restoring the physical features. In contrast, INR-based methods operate on a more compressed latent space of merely 400 dimensions, which is less than half the latent dimensionality utilized by the AutoEncoders, but they can still achieve superior performances. Besides, our SINR architecture outperforms the other INR models, demonstrating the effectiveness of the spherical filters as well as the skip connections\textcolor{black}{, and Figure \ref{fig:comp_eff} has shown that the spherical filters tend to play a much more significant role than the skip connection mechanism}. Upon observing the compelling accuracy and efficiency of the SINR structure, it becomes clear that our choice of SINR for the LAINR framework is well-justified. 

\subsubsection{Embedding the physical dynamics}
\begin{figure}
	\centering
	\includegraphics[width=.5\textwidth]{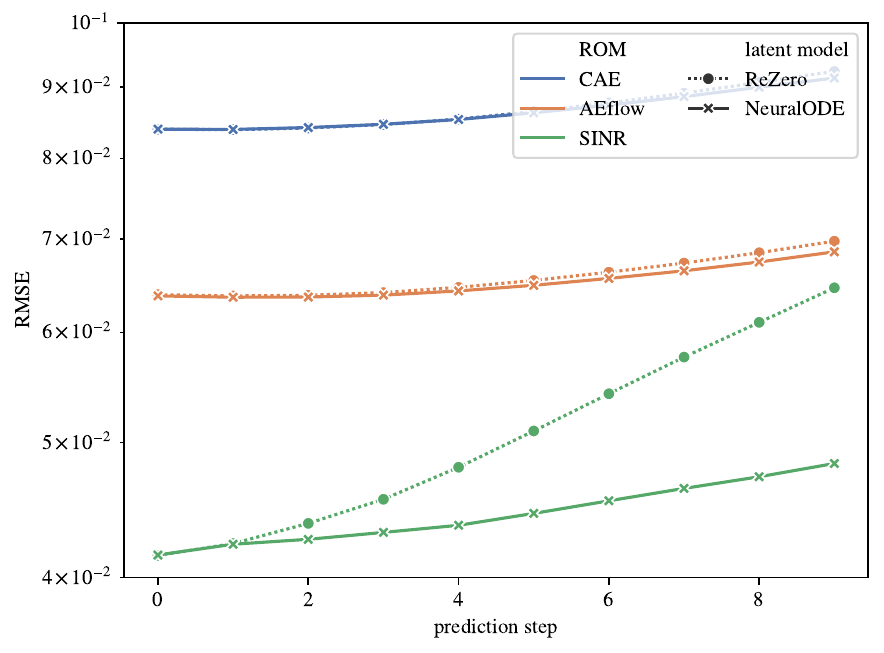}
	\caption{Comparison of the multi-step prediction errors on the physical space for the shallow-water dataset}
	\label{fig:comp-multi-step}
\end{figure}

Embedding a physical dynamical system into a latent space involves not only a precise encoding of the snapshot at each time step but also an accurate discovery of the latent dynamics, which is necessary for the assimilation process in the latent space. To evaluate both the consistency and the stability of the surrogate models for latent dynamics, the multi-step prediction RMSE
\begin{equation}
	L_{\textrm{sqrt}}^{(s)}=\frac1K\sum_{k=1}^K\mL_{\textrm{sqrt}}\left(\bm x_{k+s},\mD_{\bm\varphi}\circ\mG^s\circ\mE_{\bm\theta}(\bm x_k)\right)
\end{equation}
is utilized, where $k$ is the time index for the state $\bm x$ and $s$ indicates the number of prediction steps. We have investigated two distinct surrogate models for the latent dynamics:
\begin{itemize}
	\item \textbf{ReZero} \cite{bachlechner2021rezero}. We have followed the work of \cite{Peyron2021LAwithAE}, which implements the ReZero structure as a discrete-time dynamical model. The ReZero structure is implemented with 5 residual blocks, and the activation function is fixed as the LeakyReLU \cite{maas2013rectifier} except for the final block.
	\item \textbf{NeuralODE} \cite{chen2018NeuralODE,chen2021eventfn}. The NeuralODE model interprets the latent dynamics as a continuous-time dynamical system, where the surrogate model is implemented via a fully-connected network and trained with an RK4 integrator via the TorchDiffEq \cite{chen2018NeuralODE,chen2021eventfn} package.
\end{itemize}
We have fixed $s=1$ to define the loss function for training the latent surrogate models.
Figure \ref{fig:comp-multi-step} exhibits the comparison of the increments in prediction errors with respect to the prediction steps. Compared with CAE, AEflow shows relatively better performances probably due to its higher reconstruction accuracy as visualized in Figure \ref{fig:comp_eff}, while our SINR model presents significantly lower prediction errors. Meanwhile, by comparing the two latent surrogate models, it can be concluded that NeuralODE is more capable of capturing the latent dynamics than ReZero especially when working with SINR. In contrast, they do not show significant differences in the prediction errors when paired with the AutoEncoder baselines, which implies that their failure to obtain good latent representations probably leads to much more complicated latent dynamics.

\subsubsection{Working together with existing DA algorithms}\label{sec:compatibility}
The effectiveness of a DA algorithm is linked to the quality of the forward integrator, but an accurate surrogate model alone does not guarantee better assimilation performances, owing to the potential incompatibility between the surrogate model's non-lineality behaviors and the assimilation algorithms. To address such concern, we check the compatibility of our LAINR framework, alongside the AutoEncoder baselines, when paired with existing assimilation algorithms. 

We choose several variants of the Kalman filter method for testing, and we would like to emphasize that similar experiments can be extended to other types of assimilation algorithms including the variational methods (3/4D-Var). The ensemble size is fixed as $N=64$. Detailed implementations for the Kalman filter (KF) algorithms (EnKF, SEnKF, DEnKF, ETKF, ETKF-Q) involved in our experiments are all exhibited in \ref{tab:kalman_filter_methods}.

The observation operator $\mH:G\to S_\textrm{obs}$ is set as a random sampling on the grid $G$ of size $128\times64\times2$ as illustrated in Figure \ref{fig:regular-grid} with the number of the observations $|S_\textrm{obs}|=1024$ (6.25\% of the total number of grid points). All observations are perturbed with independent Gaussian noise of variance $(\sigma^o)^2=0.1^2$, which is approximately $10\%$ of the ground truth since the datasets are normalized to zero mean and standard deviation in advance. The initial physical background estimate is set as a perturbated observation of the ground truth via $\mH$ with a Gaussian noise $\sigma_x^b=0.1$. Additionally, the noise for the latent surrogate model is assumed as Gaussian of covariance matrix $(\sigma^m)^2\bm I_m$ without any uncertainty estimator for now, and the covariance for the uncertainty of the latent background estimate is set as $(\sigma_z^b)^2\bm I_m$.
\begin{table}
	\centering
	\caption{Configurations for assimilation on the shallow-water dataset.}
	\label{tab:sw-assconfig}
	\begin{tabular}{c|c|c}
		\Xhline{2\arrayrulewidth}
		variables    & values                                        & count \\
		\hline
		$\sigma_z^b$ & 0.01, 0.003, 0.001                            & 3     \\
		$\sigma^m$   & 0.1, 0.03, 0.01, 0.003, 0.001, 0.0003, 0.0001 & 7     \\
		inflation    & 1.02, 1.05, 1.10                              & 3     \\
		\hline
		constants    & $\sigma^o=0.1$, $\sigma_x^b=0.1$, $N=64$      & -     \\
		\Xhline{2\arrayrulewidth}
	\end{tabular}
\end{table}

\begin{figure}
	\centering
	\includegraphics[width=.8\textwidth]{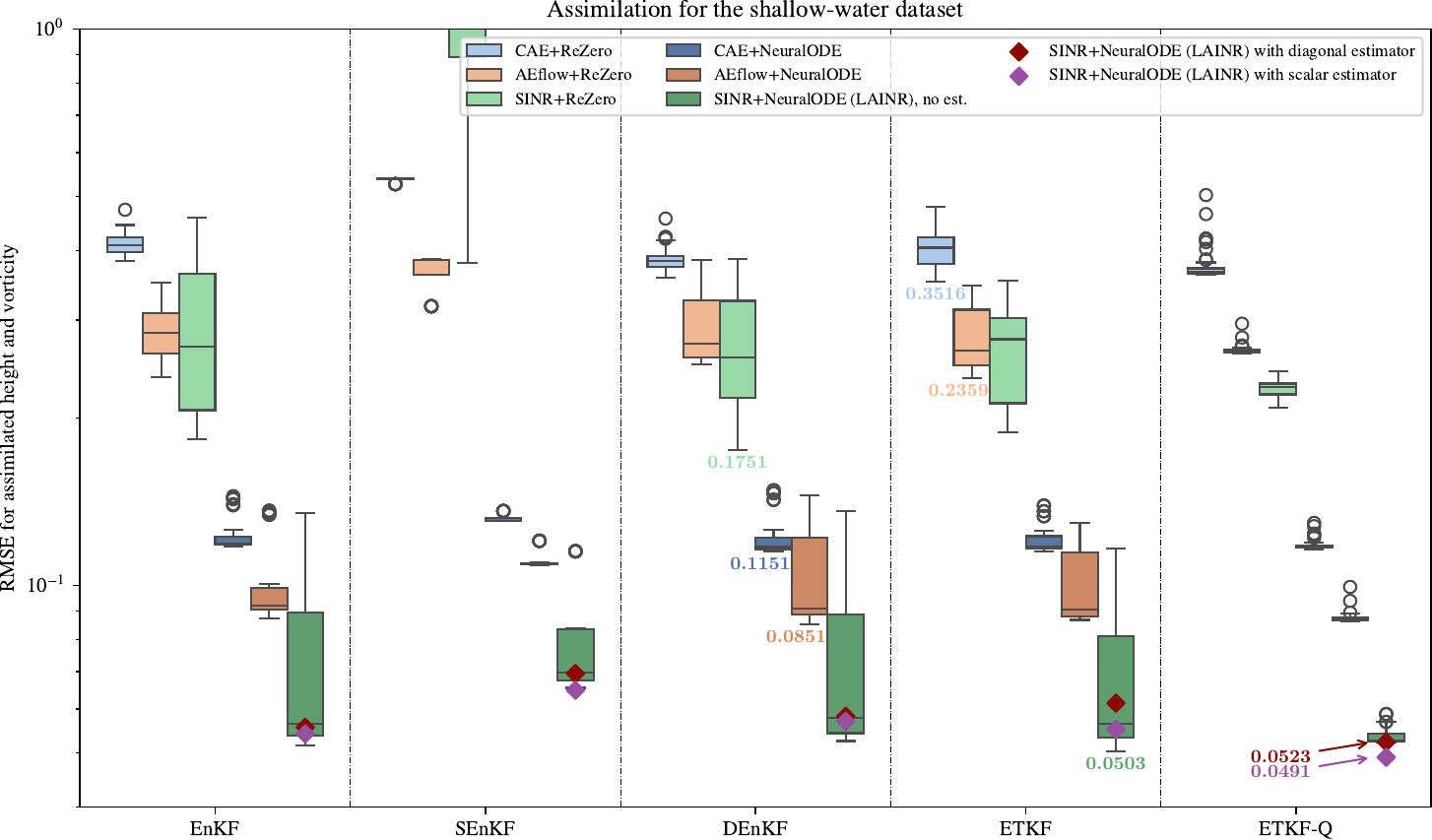}
	\caption{Comparison of assimilation performances on the shallow-water dataset.}
	\label{fig:sw_assimilation_boxfig}
\end{figure}

To work with existing DA algorithms, the latent background estimate $\sigma_z^b$, the uncertainty $\sigma^m$ of the latent models as well as the inflation needs to be specified. To study more thoroughly the behaviors of the DL models when applied to assimilation, we have conducted experiments for each model with different KF algorithms under $3\times7\times3=63$ individual configurations (Table \ref{tab:sw-assconfig}). See Figure \ref{fig:sw_assimilation_boxfig} for the assimilation performances on the shallow-water dataset, where each box contains 63 results. For each model, the best RMSE across all KFs (315 configurations in total) is marked in bold below the corresponding box.

Notably, NeuralODE significantly outperforms ReZero when combined with either the AutoEncoder baselines or our SINR model, although such huge differences have not been observed previously on the prediction tasks when evaluating the fine-tuning performances. It can be inferred that modeling the latent dynamics with a continuous-time model helps stabilize the assimilation cycle. When comparing different encoder-decoder mappings, our SINR outperforms the AutoEncoder baselines (CAE and AEflow) when working together with KFs under different assumptions on the model noise $\sigma^m$ and the latent background estimate $\sigma_z^b$. Such superiority of INR emphasizes its robustness across various configurations.
The comparison of optimal performances further validates the supremacy of the SINR model. It achieves considerably lower RMSEs than the AutoEncoder methods, reinforcing the idea that the LAINR framework is a superior choice for latent assimilation tasks. 

\subsubsection{Uncertainty estimations}\label{sec:exp-uncertainty-estimation}
To study the effectiveness of the uncertainty estimator in our LAINR framework, we have run additional experiments with the model noise given by the following estimators (Section \ref{sec:uncertainty-latent-surrogate-model}):
\begin{itemize}
	\item \textbf{scalar}: the estimator $\bm D$ is a scalar matrix with a unique trainable parameter;
	\item \textbf{diagonal}: the estimator $\bm D$ is a diagonal matrix with all of its diagonal entries being trainable.
\end{itemize}
Meanwhile, the latent background estimates are provided as described in Section \ref{sec:uncertainty-latent-background-estimate}. For each KF, the optimal performances of LAINR (SINR+NeuralODE), where only the inflation varies, combined with uncertainty estimators (abbreviated as u.e.) are marked on the box plot (Figure \ref{fig:sw_assimilation_boxfig}).

When working with different KFs, the uncertainty estimator serves as a robust method that eliminates the need for empirical tuning of the model noise level $\sigma^b$ as well as the uncertainty for the latent background estimate $\sigma_z^b$. As illustrated in Figure \ref{fig:sw_assimilation_boxfig}, the scalar estimator uniformly outperforms the diagonal estimator, and it obtains better performances than all the empirically-tuned configurations when coupled with SEnKF and ETKF-Q, while it also achieves nearly optimal results across other KFs. Such optimality may also reveal the isotropy of the LAINR prediction errors, which indicates that the SINR model has done well in the model reduction task during training. It is worth noting that the uncertainty estimator offers an effective data-driven tool to estimate the uncertainties of the trained networks, which can be particularly advantageous in situations where time or computational resources are limited, or where expert knowledge for parameter-tuning is costly or unavailable.

\subsubsection{Visualization of the assimilation processes}
\begin{figure}
	\centering
	\includegraphics[width=.95\textwidth]{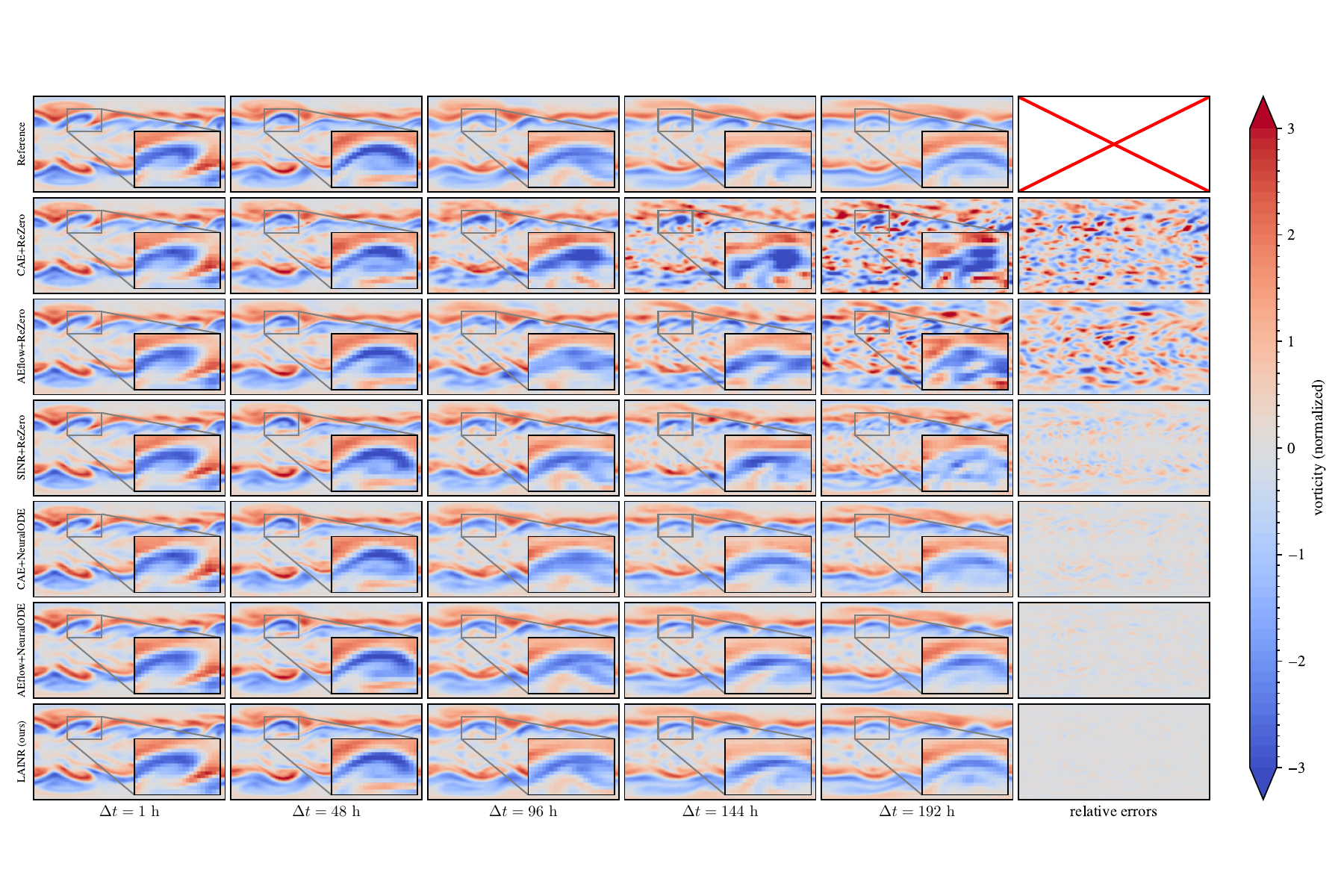}
	\caption{Assimilation vorticity fields for the shallow-water dataset}
	\label{fig:sw-ass-stamp}
\end{figure}

\begin{figure}
	\centering
	\includegraphics[width=.5\textwidth]{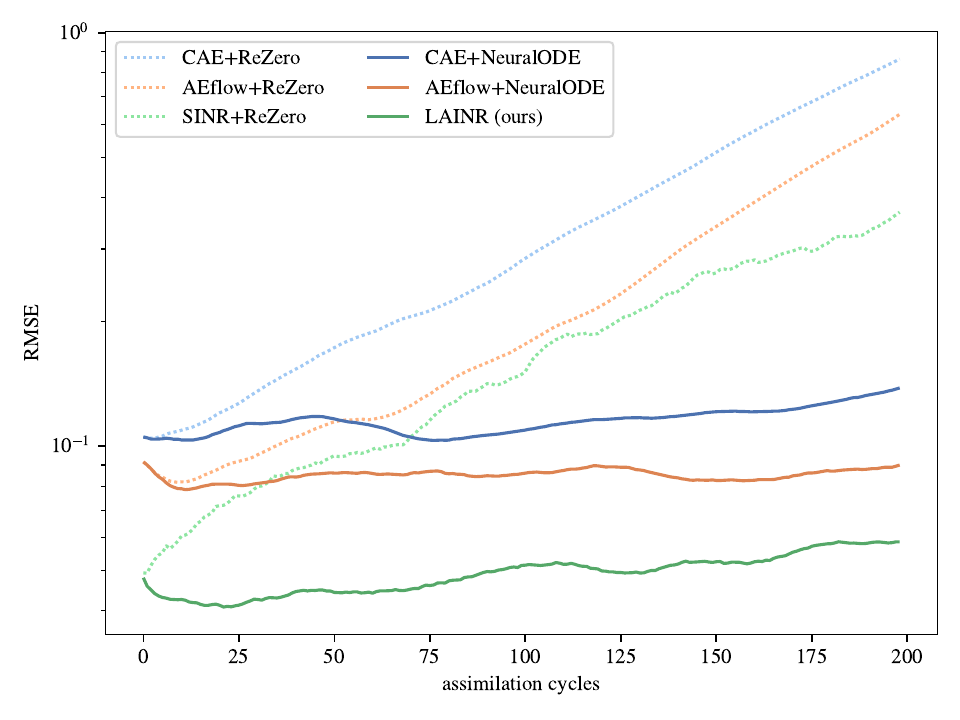}
	\caption{Increments of RMSEs during assimilation}
	\label{fig:ass-err-comp}
\end{figure}

By fixing the optimal choices of latent surrogate models, KFs as well as the assimilation configurations for each encoder-decoder mapping, we have visualized the evolution of the assimilated vorticity fields of the shallow-water dataset along with the ground truth via Figure \ref{fig:sw-ass-stamp}, where a zoom-in view for each snapshot is attached for illustration purpose. The absolute errors for the last time step based on the ground truth (reference) have been attached to the last column as well.
Figure \ref{fig:ass-err-comp} plots the detailed RMSEs for each assimilation cycle.

The AutoEncoder-based methods are unable to capture the physical dynamics during assimilation cycles when coupled with ReZero, and the assimilated states tend to diverge after 48 hours. our SINR model is able to keep tracking the dynamics until the assimilation time exceeds 144 hours (6 days) because of its high accuracy in model reduction. When using NeuralODE for the latent surrogate model, all the ROMs remain stable as the assimilation time steps increase, which does not show any divergence of errors. However, some small features of the vorticity fields have been missed by the AutoEncoders after 144 hours (6 days). Our LAINR still persistently recovers most of the physical state with relatively lower assimilation errors.
\subsubsection{Testing on a more realistic case}
To explore the capability of our proposed LAINR framework when dealing with real data, we have conducted similar comparative experiments but on a much more realistic ERA5 dataset. The study involves CAE, AEflow and our SINR as the encoder-decoder mappings. We have chosen not to advance other architectures due to their inferiority shown in the previous ablation study. Meanwhile, we have fixed the latent surrogate model as NeuralODE in that it has outperformed ReZero for both prediction and assimilation tasks, and thus we do not specify it hereafter.

\begin{figure}
	\centering
	\begin{subfigure}{.46\textwidth}
		\centering
		\vspace*{1ex}
		\includegraphics[width=\textwidth]{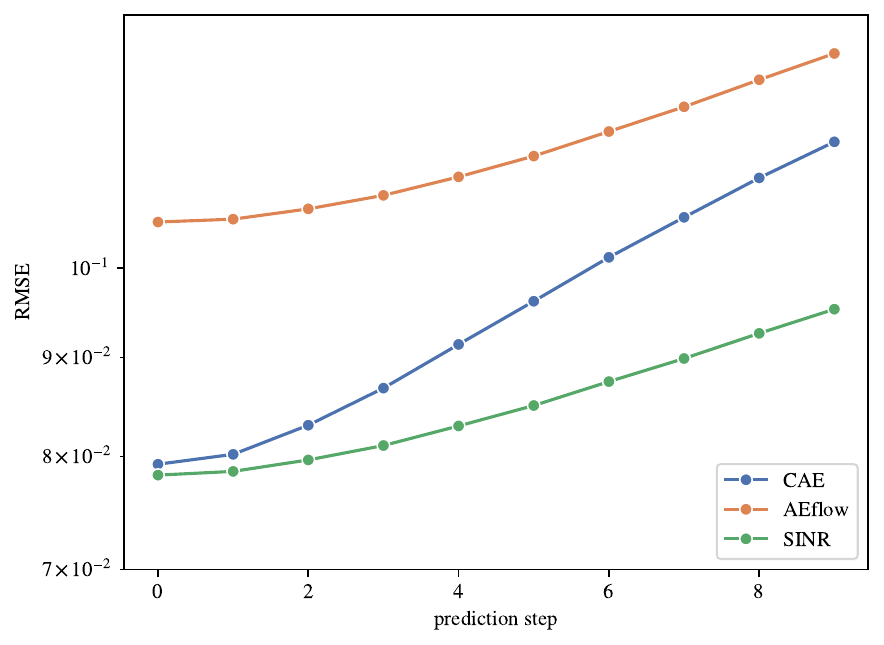}
		\caption{Multi-step prediction errors on the physical space}
		\label{fig:era5-dynamics-loss}
	\end{subfigure}
	\hfill
	\begin{subfigure}{.53\textwidth}
		\centering
		\includegraphics[width=\textwidth]{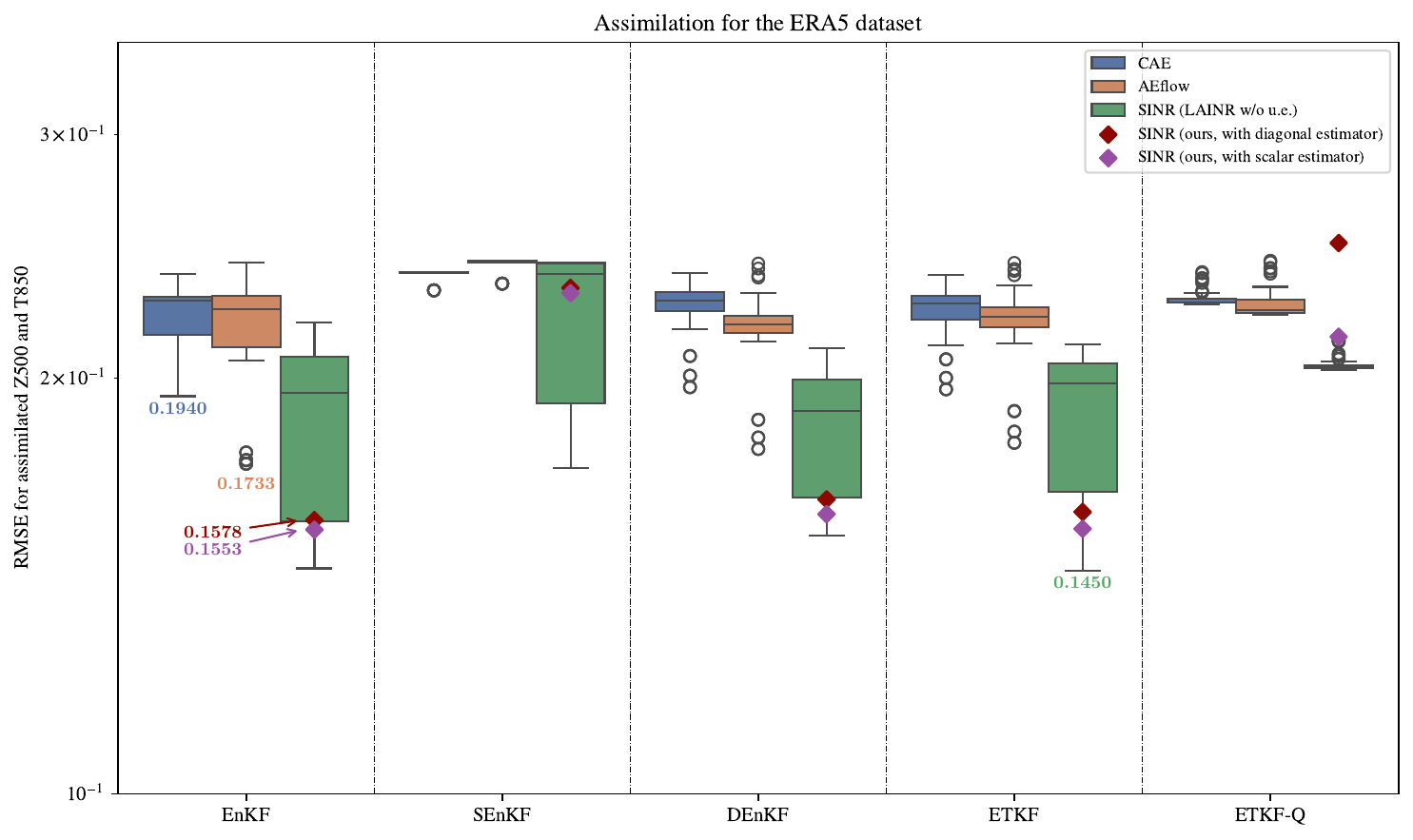}
		\vspace*{-0.2ex}
		\caption{Comparisons of RMSEs for assimilation}
		\label{fig:era5-assimilation-boxfig}
	\end{subfigure}
	\caption{Experiments on the ERA5 dataset (both truncated from above)}
\end{figure}

Same as before, our experiments start from training an encoder-decoder mapping aiming at accurately reconstructing the state variables from lower-dimensional latent representations. Then it follows that each model is used to learn the underlying latent dynamics. Figure \ref{fig:era5-dynamics-loss} displays the prediction errors (truncated from above) by computing the prediction RMSE losses on the testing dataset, and the corresponding reconstruction accuracy can be inferred by the performances at the 0-th steps.

In contrast to the shallow-water case, CAE outperforms AEflow in both the reconstruction and the prediction tasks in the ERA5 case.
Meanwhile, our LAINR still achieves the best reconstruction accuracy as well as the lowest prediction errors, which is consistent with the previous results on the shallow-water dataset. Figure \ref{fig:era5-assimilation-boxfig} exhibits the subsequent assimilation performances of the trained models when collaborating with various KFs. The configurations of the assimilation remain the same as those described in Section \ref{sec:compatibility}, which means each box contains 63 distinct parameter combinations. 

Despite AEflow's relatively poor performances in the reconstruction and the prediction tasks when compared with CAE displayed in Figure \ref{fig:era5-dynamics-loss}, it has shown comparable performances for the assimilation tasks, and it even surpasses CAE if the assimilation parameters are properly tuned (see some of the outliers in the box plot). Such a phenomenon supports the idea that higher reconstruction and prediction accuracy do not necessarily lead to better assimilation performances, and the compatibility between the surrogate models and the DA algorithms also counts. Our LAINR continues to show superior performances among all the configurations, and the uncertainty estimators turn out to be effective as well, which further strengthens the reliability of LAINR for real deployment.

\subsection{Mesh-free flexibility}\label{sec:mesh-free-flexibility}
One of the main features of our LAINR framework is its flexibility in handling inputs from partially observed or even irregular sampling sets. To demonstrate this, we have performed additional experiments on a sampling set different from what we have adopted for training.
\subsubsection{Unstructured reconstruction}
\begin{figure}
	\centering
	\includegraphics[width=.8\textwidth]{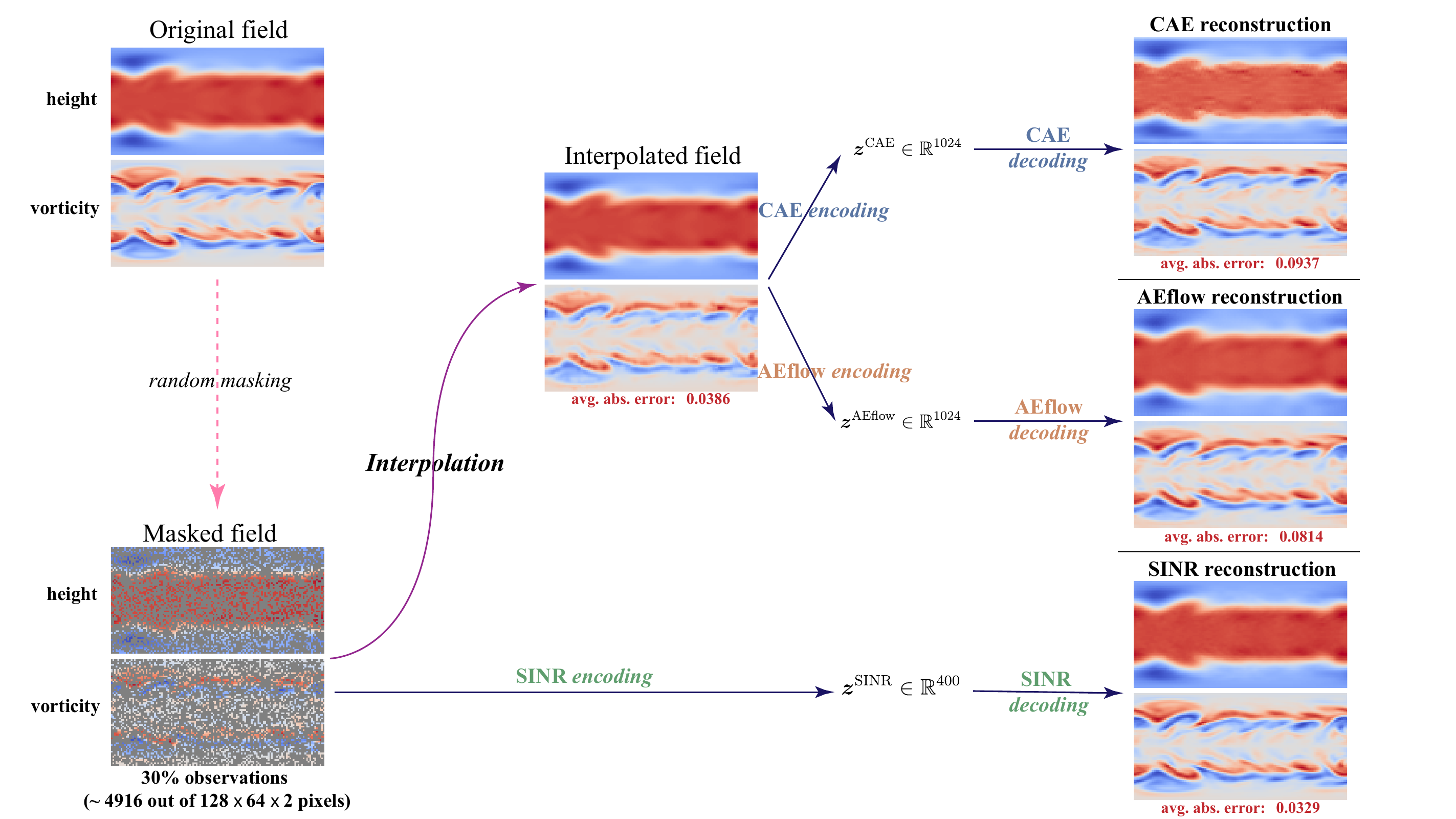}
	\caption{Reconstruction results from a random sampling set (30\% observations)}
	\label{fig:offgrid-reconstruction-30}
\end{figure}

\begin{figure}
	\centering
	\includegraphics[width=.8\textwidth]{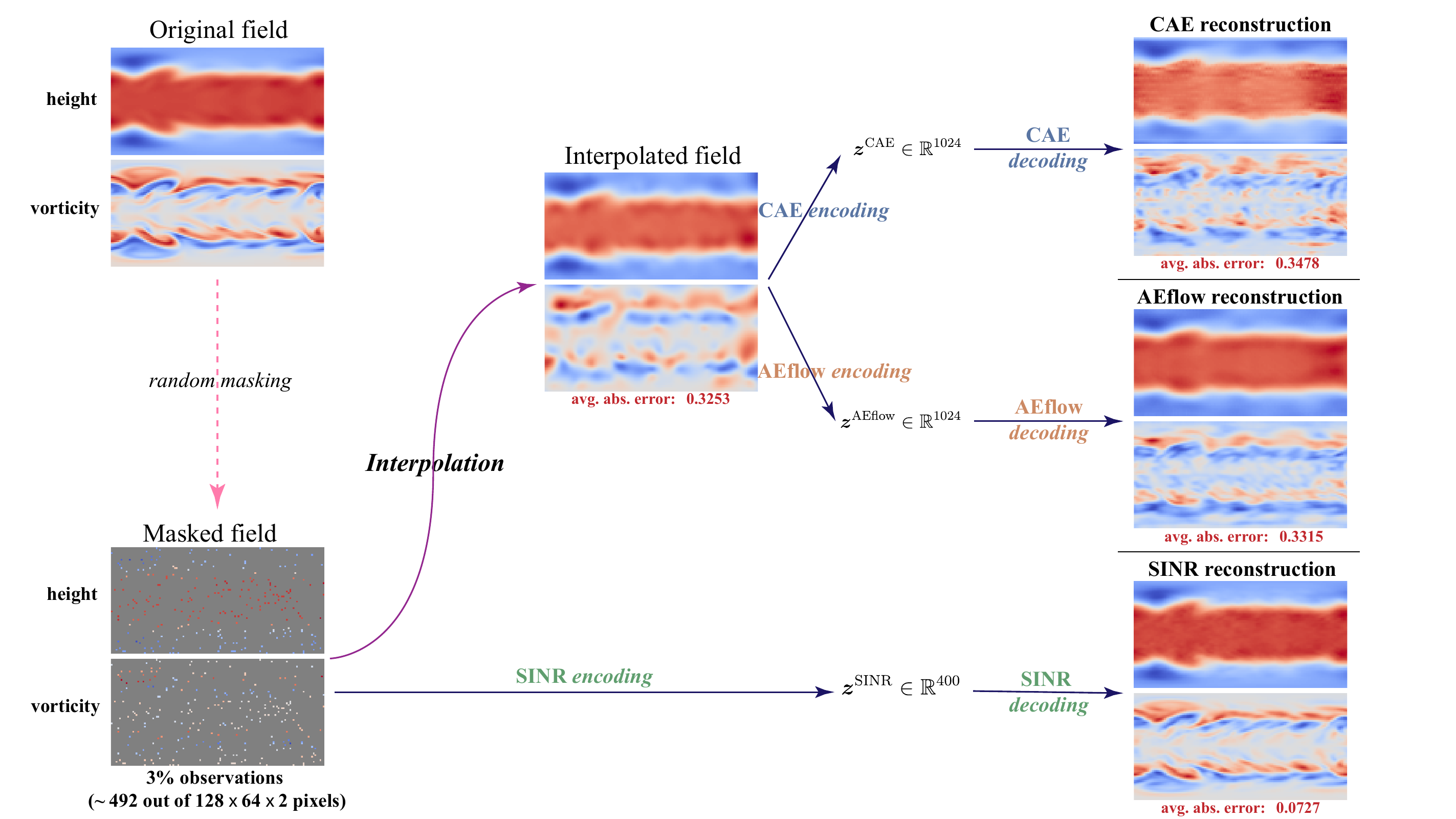}
	\caption{Reconstruction results from a random sampling set (3\% observations)}
	\label{fig:offgrid-reconstruction-3}
\end{figure}

To illustrate the effectiveness of our LAINR framework for unstructured cases, we have picked a snapshot among the testing data of the shallow-water dataset, and then randomly masked out a subset from the collection of the grid points to provide unstructured observations. For the AutoEncoder-based models (CAE and AEflow), an interpolated field is first computed and fed into as the input, where the interpolation is implemented via an RBF interpolator from the SciPy package \cite{SciPy} with the cubic kernel\footnote{\url{https://docs.scipy.org/doc/scipy/reference/generated/scipy.interpolate.RBFInterpolator.html}}. After that, an encoding-decoding process is performed to evaluate the accuracy of the latent representations $\bm z^{\textrm{CAE}}$ and $\bm z^{\textrm{AEflow}}$ in order to emulate what an AutoEncoder-based model will probably do. On the contrary, the masked field can be directly sent to the SINR model for encoding, and the only difference from the gridded case is that the number of the summation terms of \eqref{eq:INR-minimization} is reduced. The accuracy of the latent representation $\bm z^{\mathrm{SINR}}$ is also implied by the reconstruction error after the decoding. See Figure \ref{fig:offgrid-reconstruction-30} for the comparisons of the reconstruction results with 30\% random observations (few) and Figure \ref{fig:offgrid-reconstruction-3} with 3\% observations (rare).

Under different observation ratios, the AEflow model performs slightly better than CAE, but both of the AutoEncoder baselines have failed to reconstruct the masked field accurately, especially when the observations are limited. On the contrary, our SINR model has shown consistent superiority, and it is still able to recover the system states with rare observations, where nearly all of the physical features have been masked out.

\begin{figure}
	\centering
	\includegraphics[width=.85\textwidth]{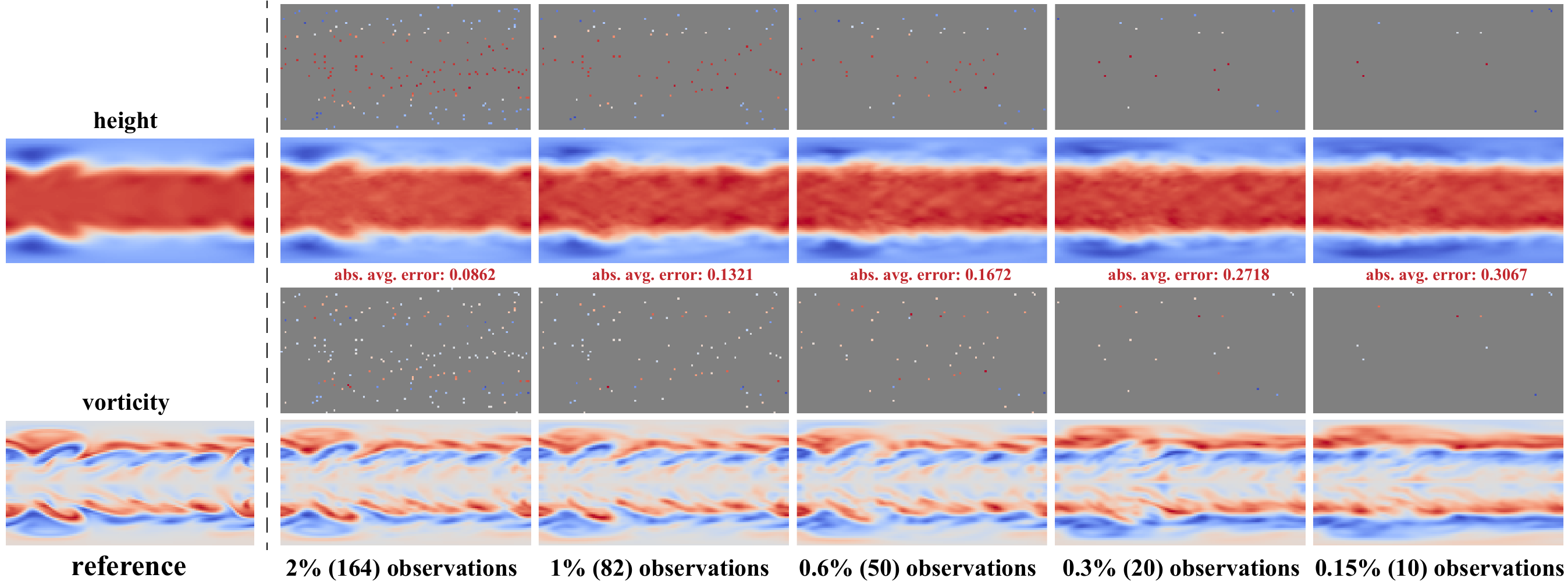}
	\caption{Explore the limit of SINR for rare observations}
	\label{fig:offgrid-reconstruction-limit}
\end{figure}

In addition, we have pushed our SINR model to the limit when the observations are extremely rare. We have conducted similar experiments with 2\%, 1\%, 50, 20 and 10 observations for each feature respectively, and the results are shown in Figure \ref{fig:offgrid-reconstruction-limit}. To our surprise, the SINR model can still recover the general fluid structures before the number of observations drops to 20 or less, which is only approximately 0.3\% of the total number of grid points. This is strong evidence of the robustness of our SINR model for unstructured observations, and it also demonstrates the potential of our LAINR framework for real-world applications where the observations usually fail to be complete or structured.
\subsubsection{Zero-shot assimilation}
Previous experiments are all based on the same regular latitude-longitude grid for both training and assimilation to make comparisons with AutoEncoder baselines. To emphasize the flexibility of our proposed LAINR framework for different sampling sets as inputs, we have performed the same assimilation procedures but on a staggered grid $G'$ as illustrated in Figure \ref{fig:staggered-grid}. Once the SINR model together with NeuralODE has been well-trained on the grid $G$, we apply LAINR to assimilate the observations on the grid $G'$. Note that under such circumstances, \underline{no observations from $G$} used for training are provided, and only random observations from a subset of $G'$ ($|S_\mathrm{obs}'|=1024$) are available for the assimilation process. Here, the grid $G$ for training and the grid $G'$ do not intersect, and we refer to such setting as the \textit{zero-shot assimilation}.

\begin{figure}
	\centering
	\includegraphics[width=.5\textwidth]{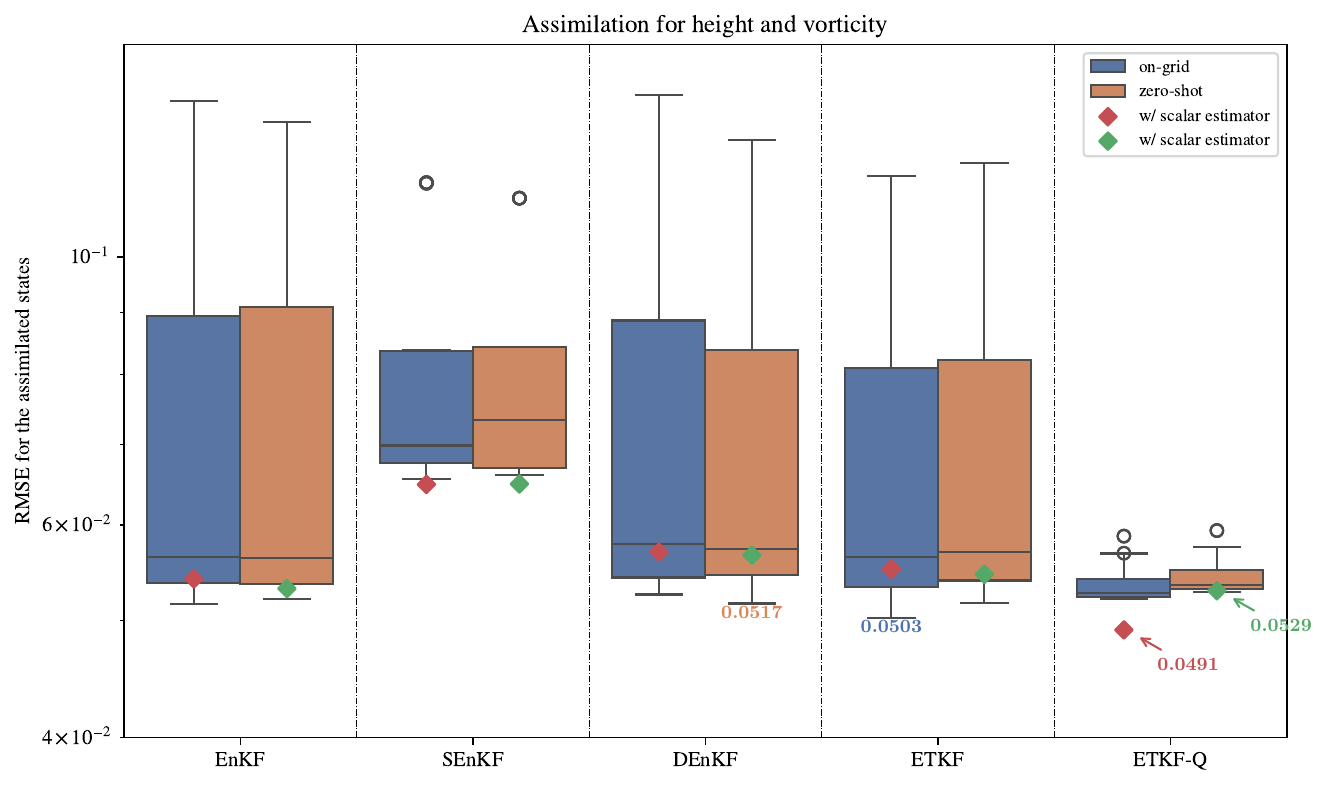}
	\caption{Comparisons of RMSEs for zero-shot assimilation on the staggered grid}
	\label{fig:zero-shot-boxplot}
\end{figure}

The parameter configurations of the assimilation remain the same as those of Section \ref{sec:compatibility}.
See Figure \ref{fig:zero-shot-boxplot} for the on-grid assimilation (on $G$, copied from Figure \ref{fig:sw_assimilation_boxfig}) and the zero-shot assimilation (on $G'$). As shown in the box plot, when working with various KFs, the zero-shot performance of our LAINR framework is comparable to the on-grid performance, indicating the robustness of LAINR on zero-shot tasks when inconsistent spatially irregular observations may appear.


\section{Conclusion and future work}\label{sec:conclusion}
In this study, we have proposed a latent assimilation framework called Latent Assimilation with Implicit Neural Representations (LAINR), which offers a new integration of machine learning techniques and data assimilation concepts. Besides, our LAINR framework opens up new possibilities for processing and assimilating complex data systems by using the reduce-order technique. Apart from the theoretical analysis of LAINR, in comparison to existing AutoEncoder-based models, experiments have shown that LAINR surpasses the previous models in terms of practical performance.

The main features of our LAINR framework can be summarized as follows. Same as previous AutoEncoder-based models, LAINR's capability for non-linear embedding introduces the possibility that the complex dynamics of physical states can be transformed into simpler latent dynamics for learning, but the difference lies in the fact that LAINR exhibits a clear superiority in scalability. Since the architecture of LAINR uses coordinates as inputs, it effortlessly accommodates increasing dimensionality and resolution unlike most classical models, which can hardly avoid re-designing the whole structure due to their reliance on the input structures.

One of the biggest advantages of LAINR is perhaps its flexibility in handling inputs from partially observed or ungridded data, which becomes more suitable under potential hardware limitations when complete observations are not always guaranteed such as satellite or radar observations.
Unlike the AutoEncoder baselines treating the state field as a structured grid of state features, LAINR stores the entire state field as a continuous mapping and such a special perspective provides a more accurate description of physical features, aiding in the learning of model-reduction mappings.
Meanwhile, the availability of a continuous field may help establish more complicated observation operators such as the solution of radiative-transfer models in that the interpolation is no longer necessary.
As a consequence, its utility scope is wider in real-world scenarios compared with AutoEncoder-based architectures.

Nonetheless,
it is also important to acknowledge that areas where further research and development are needed, including the testing of LAINR on real observation datasets to provide further validations of its effectiveness and robustness. Besides, future work should explore more advanced uncertainty estimators as well. The SINR presented in the LAINR framework acts as a novel approach to construct an INR mapping from integer-indexed basis functions, and a similar idea can be applied to other geometrical structures other than the 2D sphere. Finding a suitable collection of orthogonal basis functions for general Riemann manifolds to obtain the convergence guarantee may become a future direction.

\section*{Reproducibility}
All the codes for our proposed LAINR framework and the corresponding experiments are available online at GitHub (\url{https://github.com/zylipku/LAINR}) including the data generation for the shallow-water datasets and the full list of assimilation results.
\section*{Acknowledgments}
Zhuoyuan Li and Pingwen Zhang are supported in part by the National Natural Science Foundation of China (No. 12288101). Bin Dong is supported in part by the New Cornerstone Investigator Program.


\bibliographystyle{spmpsci} 

\bibliography{LAINR-refs}


\appendix

\section{An Alternative for the SINR representability}\label{sec:SINR-representability-revisited}
In Proposition \ref{prop:SINR-representability}, we have deduced a subset $\mathcal{S}_{T}$ of the set of all possible SINRs, $\mathcal{S}_{\mathrm{SINR}}$. However, we have to point out that there exist many more functions excluded by $\mathcal{S}_T$ that can be represented by a SINR.

Denote by $\bm D^{(l)}=\diag\left\{d_1^{(l)},\cdots,d_h^{(l)}\right\}$ the diagonal matrix with entries $\left(d_1^{(l)},\cdots,d_h^{(l)}\right)^\trans=\bm g_l(\bm p)$ as defined in \eqref{eq:SINR-updates}. Consequently, the update formulas \eqref{eq:SINR-updates} can be transformed into the following
\begin{equation}\label{eq:SINR-updates-linear-algebra-version}
	\begin{aligned}
		\bm \gamma^{(0)} & =\bm D^{(0)}\bm 1,                                                                       \\
		\bm \gamma^{(l)} & =\bm D^{(l)}\left(\bm W^{(l)}\bm \gamma^{(l-1)}+\bm b^{(l)}\right),\quad l=1,2,\cdots,L, \\
		\bm f(\bm p)     & =\sum_{l=0}^L\left(\tilde{\bm W}^{(l)}\bm \gamma^{(l)}+\tilde{\bm b}^{(l)}\right).
	\end{aligned}
\end{equation}
By using basic linear algebra, we can expand the update formulas \eqref{eq:SINR-updates-linear-algebra-version} into
\begin{equation}\label{eq:SINR-updates-linear-algebra-version-expanded}
	\begin{aligned}
		\bm \gamma^{(0)} & =\bm D^{(0)}\bm 1,                                                                                                                                              \\
		\bm \gamma^{(1)} & =\bm D^{(1)}\bm W^{(1)}\bm D^{(0)}\bm 1+\bm D^{(1)}\bm b^{(1)},                                                                                                 \\
		\bm \gamma^{(2)} & =\bm D^{(2)}\bm W^{(2)}\bm D^{(1)}\bm W^{(1)}\bm D^{(0)}\bm 1+\bm D^{(2)}\bm W^{(2)}\bm D^{(1)}\bm b^{(1)}+\bm D^{(2)}\bm b^{(2)},                              \\
		                 & \vdots                                                                                                                                                          \\
		\bm \gamma^{(L)} & =\bm D^{(L)}\bm W^{(L)}\bm D^{(L-1)}\cdots\bm W^{(1)}\bm D^{(0)}\bm 1+\sum_{l=1}^L\bm D^{(L)}\bm W^{(L)}\cdots\bm D^{(l+1)}\bm W^{(l+1)}\bm D^{(l)}\bm b^{(l)}, \\
		\bm f(\bm p)     & =\sum_{l=0}^L\left(\tilde{\bm W}^{(l)}\bm \gamma^{(l)}+\tilde{\bm b}^{(l)}\right).
	\end{aligned}
\end{equation}
The proof of Proposition \ref{prop:SINR-representability} essentially sets the weight matrix $\bm W^{(l)}$ and the output bias $\tilde{\bm b}^{(l)}$ for each layer $l$ as zero to obtain a subset $\mathcal{S}_T$ easy for analysis, which is too rough to study the representability of the SINR architecture. To dive a bit more deeply into the representability, here we consider an alternative setting, where $\bm W^{(l)}$ is set as a matrix with its unique non-zero $(i_l,i_{l-1})$-th entry as 1, and all the bias terms are set to zero. Then
\begin{equation}
	\begin{aligned}
		\gamma_k^{(0)} & =d_k^{(0)},                                                          \\
		\gamma_k^{(1)} & =\delta_{k,i_1}d_{i_1}^{(1)}d_{i_0}^{(0)},                           \\
		\gamma_k^{(2)} & =\delta_{k,i_2}d_{i_2}^{(2)}d_{i_1}^{(1)}d_{i_0}^{(0)},              \\
		               & \vdots                                                               \\
		\gamma_k^{(L)} & =\delta_{k,i_L}d_{i_L}^{(L)}d_{i_{L-1}}^{(L-1)}\cdots d_{i_0}^{(0)},
	\end{aligned}
\end{equation}
where $k$ is the index of the $k$-th coordinates of these vectors, and $\delta$ stands for the Dirac delta function. It follows that each coordinate of the output $\bm f(\bm p)$ can be written as a linear combination of the graded spherical harmonics
\begin{equation}
	f_k(\bm p)\in\Span\left(d_{i_0}^{(0)}, d_{i_1}^{(1)}d_{i_0}^{(0)},\cdots, d_{i_L}^{(L)}d_{i_{L-1}}^{(L-1)}\cdots d_{i_0}^{(0)}\right),
\end{equation}
where each $d_{i_l}^{(l)}$ is a linear combination of the spherical harmonics $Y^{-D}_{D+l}(\bm p)$, $Y^{-(D-1)}_{D-1+l}(\bm p)$, $\cdots$, $Y^0_{0+l}(\bm p)$, $\cdots$, $Y^{D-1}_{D-1+l}(\bm p)$, $Y^D_{D+l}(\bm p)$. Specially, if one has picked a linear combination $d_{i_l}^{(l)}$ for each layer $l$, then their multiplication $d_{i_L}^{(L)}d_{i_{L-1}}^{(L-1)}\cdots d_{i_0}^{(0)}$ can always be represented by a SINR.

\section{Detailed dataset configurations}\label{sec:dataset_configs}
\subsection{Shallow-water equations}\label{sec:sw_configs}
The spherical shallow-water equation on the 2D sphere writes
\begin{equation}
	\begin{aligned}
		\frac{\md\bm u}{\md t} & =-f\bm k\times\bm u-g\nabla h+\nu\Delta\bm u, \\
		\frac{\md h}{\md t}    & =-h\nabla\cdot\bm u+\nu\Delta h,
	\end{aligned}
\end{equation}
where $\bm u=(u,v)^\trans$ is the 2D velocity field tangent to the sphere surface, and $h$ is the thickness of the fluid layer, both of which are defined on a channel domain $(\lambda,\phi)\in[0,2\pi]\times[-\pi/2,\pi/2]$. The Coriolis parameter $f=2\Omega\sin\phi$ with the angular velocity $\Omega$ and the gravity $g$ are fixed in consistency with the real earth surface.

The model is set up by initializing two symmetric zonal flows representing typical mid-latitude tropospheric jets
\begin{equation}
	v|_{t=0}=0,\quad u|_{t=0}=
	\begin{cases}
		\frac{u_m}{e_n}\exp\left(\frac1{(\phi-\phi_0)(\phi-\phi_1)}\right), & \phi_0<|\phi|<\phi_1, \\
		0                                                                   & \textrm{otherwise},
	\end{cases}
\end{equation}
which is slightly modified from \cite{Galewsky-2004} to create symmetric fields, where $\phi_0=\pi/7$ and $\phi_1=\pi/2-\phi_0$ gives the boundary of the jet. $e_n=\exp(-4/(\phi_1-\phi_0)^2)$ is the normalizer so that the maximal zonal wind $u_m$ lies at the jet's mid-point. The thickness $h$ is initialized as a balanced height field, which is obtained by integrating the balance equation
\[gh(\phi)=gh_0-\int^\phi au(\phi')\left(f+a^{-1}u(\phi')\tan\phi'\right)\md \phi'.\]
The constant $h_0$ is chosen so that the global mean of $h$ is $10^4$. The balanced field is then perturbed by adding a localized bump
\begin{equation}
	h'(\lambda,\phi)=\hat h\exp\left[-(\lambda/\alpha)^2-((\phi_2-\phi)/\beta)^2\right]\cos\phi,
\end{equation}
where $\hat h=120$, $\alpha=1/3$, $\beta=1/15$, and $\phi_2=\pi/4$.

The datasets for both training and testing are generated via a sample script from the Dedalus project \cite{dedalus} based on Python language within a spectral framework. The maximum velocity $u_m$ is randomly sampled from the uniform distribution on the interval $(60,80)$ to generate distinct 20 trajectories for separation of training and testing. Each trajectory has been recorded from the 360-th hour to the 600-th hour with a 1-hour interval to capture rich physical features, and the spatial resolution is set as $128\times64$. The vorticity $w=\nabla\times\bm u$ and the thickness $h$ are the two features for network training and assimilation.
\subsection{ERA5 dataset}\label{sec:era5_configs}
We have made use of the WeatherBench \cite{Rasp2020WeatherBench} dataset, which is a resampled version of the ERA5 dataset specially designed for DL training. We select the subset of latitude-longitude resolution $2.8125^\circ\times2.8125^\circ$, corresponding to exactly a spatial resolution of $128\times64$. The Z500 and T850 fields from 2006 to 2015 (10 years) with a 1-hour interval make up the training dataset, and the testing dataset is composed of the fields from 2017 and 2018.
\section{AutoEncoder baselines}\label{sec:ae_baselines}
\subsection{Detailed network structures}\label{sec:net-structures}
The detailed propagations for both the CAE and AEflow structures are listed in Table \ref{tab:cae_prop} and Table \ref{tab:aeflow_prop}, respectively, where the batch size is replaced by ``$-1$''. As for the activation functions, each convolutional layer is followed by a LeakyReLU \cite{maas2013rectifier} with a 0.2 slope except the last output layers for the two structures.
\begin{table}[b]
	\centering
	\caption{Forward propagation for the CAE baseline}
	\begin{tabular}{c|c|c}
		\Xhline{3\arrayrulewidth}
		                         & \textbf{Layer}       & \textbf{Output Shape} \\
		\Xhline{2\arrayrulewidth}
		\multirow{5}{*}{Encoder} & Input                & $(-1, 2, 128, 64)$    \\
		\cline{2-3}
		                         & ConvBlock-1          & $(-1, C, 64, 32)$     \\
		\cline{2-3}
		                         & ConvBlock-2          & $(-1, C, 32, 16)$     \\
		\cline{2-3}
		                         & ConvBlock-3          & $(-1, 8, 16, 8)$      \\
		\hline
		\multirow{5}{*}{Decoder} & Input                & $(-1, 8, 16, 8)$      \\
		\cline{2-3}
		                         & ConvTransposeBlock-1 & $(-1, C, 32, 16)$     \\
		\cline{2-3}
		                         & ConvTransposeBlock-2 & $(-1, C, 64, 32)$     \\
		\cline{2-3}
		                         & ConvTransposeBlock-3 & $(-1, 2, 128, 64)$    \\
		\Xhline{3\arrayrulewidth}
	\end{tabular}
	\label{tab:cae_prop}
\end{table}
\begin{table}
	\centering
	\caption{Forward propagation for the AEflow baseline}
	\begin{tabular}{c|c|c}
		\Xhline{3\arrayrulewidth}
		                          & \textbf{Layer}                  & \textbf{Output Shape} \\
		\Xhline{2\arrayrulewidth}
		\multirow{11}{*}{Encoder} & Input                           & $(-1, 2, 128, 64)$    \\
		\cline{2-3}
		                          & Conv2d-1                        & $(-1, 4, 128, 64)$    \\
		\cline{2-3}
		                          & ResBlock-1                      & $(-1, 4, 128, 64)$    \\
		\cline{2-3}
		                          & $\cdots\cdots$                  & $(-1, 4, 128, 64)$    \\
		\cline{2-3}
		                          & ResBlock-{$N_{\textrm{res}}$}   & $(-1, 4, 128, 64)$    \\
		\cline{2-3}
		                          & Conv2d-2                        & $(-1, 4, 128, 64)$    \\
		\cline{2-3}
		                          & Skip connection from ResBlock-1 & $(-1, 4, 128, 64)$    \\
		\cline{2-3}
		                          & ConvBlock-1                     & $(-1, 8, 64, 32)$     \\
		\cline{2-3}
		                          & ConvBlock-2                     & $(-1, 16, 32, 16)$    \\
		\cline{2-3}
		                          & ConvBlock-3                     & $(-1, 16, 16, 8)$     \\
		\cline{2-3}
		                          & Conv2d-3                        & $(-1, 8, 16, 8)$      \\
		\hline
		\multirow{11}{*}{Decoder} & Input                           & $(-1, 8, 16, 8)$      \\
		\cline{2-3}
		                          & Conv2d-1                        & $(-1, 16, 16, 8)$     \\
		\cline{2-3}
		                          & ConvTransposeBlock-1            & $(-1, 16, 32, 16)$    \\
		\cline{2-3}
		                          & ConvTransposeBlock-2            & $(-1, 8, 64, 32)$     \\
		\cline{2-3}
		                          & ConvTransposeBlock-3            & $(-1, 4, 128, 64)$    \\
		\cline{2-3}
		                          & ResBlock-1                      & $(-1, 4, 128, 64)$    \\
		\cline{2-3}
		                          & $\cdots\cdots$                  & $(-1, 4, 128, 64)$    \\
		\cline{2-3}
		                          & ResBlock-{$N_{\textrm{res}}$}   & $(-1, 4, 128, 64)$    \\
		\cline{2-3}
		                          & Conv2d-2                        & $(-1, 4, 128, 64)$    \\
		\cline{2-3}
		                          & Skip connection from ResBlock-1 & $(-1, 4, 128, 64)$    \\
		\cline{2-3}
		                          & Conv2d-3                        & $(-1, 2, 128, 64)$    \\
		\Xhline{3\arrayrulewidth}
	\end{tabular}
	\label{tab:aeflow_prop}
\end{table}
\subsection{Grid search}\label{sec:grid-search}
As mentioned in Section \ref{sec:model-configs}, we have conducted a grid search on the configurations for the network structures and then fixed the optimal parameters to make a fair comparison with our proposed framework. The grid search results for the CAE and AEflow structures based on the shallow-water dataset are summarized in Table \ref{tab:ablation}. We fix the learning rate as $10^{-3}$ except for the starred results due to training divergence, where we instead set the learning rate as $10^{-4}$. All the networks have been trained with 30000 epochs to avoid underfitting, and the optimal configurations are highlighted in boldface.
\begin{table}
	\centering
	\caption{Testing RMSEs for different configurations of AutoEncoder baselines (lower is better).}
	\begin{tabular}{c|c|c|c|c}
		\Xhline{3\arrayrulewidth}
		AutoEncoder             &                       & $K=3$  & $K=5$                       & $K=7$         \\
		\Xhline{2\arrayrulewidth}
		\multirow{3}{*}{CAE}    & $C=16$                & 0.1215 & 0.1114                      & 0.1044        \\
		\cline{2-5}
		                        & $C=32$                & 0.1044 & \underline{\textbf{0.0841}} & 0.0979        \\
		\cline{2-5}
		                        & $C=64$                & 0.1016 & $0.1231^\ast$               & $0.1060^\ast$ \\
		\Xhline{2\arrayrulewidth}
		\multirow{3}{*}{AEflow} & $N_{\textrm{res}}=4$  & 0.0944 & \underline{\textbf{0.0782}} & $0.1404^\ast$ \\
		\cline{2-5}
		                        & $N_{\textrm{res}}=8$  & 0.0839 & 0.0937                      & 0.1080        \\
		\cline{2-5}
		                        & $N_{\textrm{res}}=12$ & 0.0938 & 0.1024                      & $0.1390^\ast$ \\
		\Xhline{3\arrayrulewidth}
	\end{tabular}
	\label{tab:ablation}
\end{table}

\section{Algorithms for associated Kalman filter methods}\label{tab:kalman_filter_methods}
\subsection{EnKF}
The basic Ensenble Kalman Filter (EnKF)
\subsubsection{Forecast step}
The perturbed forward propagation writes
\begin{equation}
	x_k^{b,j}=\mathcal{M}_k\left(x_{k-1}^{a,j}\right)+\varepsilon_k^{M,j},\quad \varepsilon_k^{M,j}\sim\mathcal{N}\left(0,\Sigma_k^M\right).
\end{equation}
\subsubsection{Analysis step}
\begin{itemize}
	\item Obtain the ensemble mean and anomaly matrix from the ensembles:
	      \begin{equation}
		      \begin{aligned}
			      x_k^b & = \frac1N\sum_{j=1}^Nx_k^{b,j},                                                      \\
			      X_k^b & = \frac1{\sqrt{N-1}}\left(x_k^{b,j} - x_k^b\right)_{j=1}^N\in\mathbb{R}^{n\times N},
		      \end{aligned}
	      \end{equation}

	      where the background estimate for the covariance matrix is described as $P_k^b=X_k^bX_k^{b\trans}$.
	\item Sample and calculate the perturbed observation vectors and the corresponding innovation vectors:
	      \begin{equation}
		      \begin{aligned}
			      z_k^{o,j} & = y_k^o + \varepsilon_k^{o,j},\quad\varepsilon_k^{o,j}\sim\mathcal{N}\left(0,\Sigma_k^o\right); \\
			      d_k^j     & = z_k^{o,j} - \mathcal{H}_k\left(x_k^{b,j}\right).
		      \end{aligned}
	      \end{equation}
	\item Evaluate the Kalman gain matrix
	      \begin{equation}
		      \begin{aligned}
			      K_k & = P_k^bH_k^\trans \left(H_kP_k^bH_k^\trans +\Sigma_k^o\right)^{-1}                           \\
			          & = X_k^bX_k^{b\trans}H_k^\trans \left(H_kX_k^bX_k^{b\trans}H_k^\trans +\Sigma_k^o\right)^{-1} \\
			          & = P_{xy}\left(P_{yy}+\Sigma_k^o\right)^{-1}.
		      \end{aligned}
	      \end{equation}
	      Here, $P_{xy}$ and $P_{yy}$ are given by
	      \begin{equation}
		      P_{xy}=X_k^bY_k^{b\trans},\quad P_{yy}=Y_k^bY_k^{b\trans},
	      \end{equation}
	      where $Y_k^b$ is the anomaly matrix of $\mH_k\left(x_k^{b,j}\right)$.
	\item Update the analyzed ensembles:
	      \begin{equation}
		      x_k^{a,j}=x_k^{b,j}+K_kd_k^j.
	      \end{equation}
\end{itemize}
\subsection{SEnKF}
Stochastic EnKF (Algorithm 6.3 in \cite{Bocquet2016})

Same as the previous EnKF, except that the Kalman gain matrix is implemented stochastically as
\begin{equation}
	K_k=X_k^bY_k^{p\trans}\left(Y_k^pY_k^{p\trans}\right)^{-1},
\end{equation}
where $Y_k^p$ is the anomaly matrix of $y_k^{p,j}=\mH_k\left(x_k^{b,j}\right)-\varepsilon_k^{o,j}$.

\subsection{ETKF}
ETKF (Algorithm 6.4 in \cite{Bocquet2016})

\subsubsection{Forecast step}
\begin{itemize}
	\item Forward propagation:
	      \begin{equation}
		      x_k^{b,j}=\mathcal{M}_k\left(x_{k-1}^{a,j}\right)+\varepsilon_k^{M,j},\quad \varepsilon_k^{M,j}\sim\mathcal{N}\left(0,\Sigma_k^M\right);
	      \end{equation}
	      perturbed propagation
\end{itemize}

\subsubsection{Analysis step}
\begin{itemize}
	\item Obtain the ensemble means $x_k^b$ and the anomaly matrix $X_k^b$ from the state ensembles $x_k^{b,j}$ and $y_k^b$ and the anomaly matrix $Y_k^b$ from $y_k^{b,j}=\mH\left(x_k^{b,j}\right)$, respectively as before.
	\item Calculate the transforming matrix
	      \begin{equation}
		      T_k = \left(I_N+Y_k^{b\trans}\left(\Sigma_k^o\right)^{-1}Y_k^b\right)^{-1} = \left(I_N + S_k^\trans S_k\right)^{-1}
	      \end{equation}
	      for $S_k=\left(\Sigma_k^o\right)^{-1/2}Y_k^b$;
	\item Update the analyzed ensembles with the innovation vector $d_k=y_k^o-y_k^b$:
	      \begin{equation}
		      \begin{aligned}
			      w_k^a     & = \left(I_N + Y_k^{b\trans}\left(\Sigma_k^o\right)^{-1} Y_k^b\right)^{-1}Y^{b\trans}\left(\Sigma_k^o\right)^{-1}\left(y_k^o-y_k^b\right)=T_k\left(\Sigma_k^o\right)^{-1}d_k \\
			      x_k^{a,j} & = x_k^a + \sqrt{N-1}X_k^a[:, j],
		      \end{aligned}
	      \end{equation}
	      where $U$ is a random orthogonal matrix such that $U1=1$.
\end{itemize}

\subsection{ETKF-Q}
The ETKF-Q algorithm borrowed from \cite{Peyron2021LAwithAE} (a special case of IEnKS-Q \cite{Fillion2020IEnKS}).

\subsubsection{Initialization}

Construct $U\in\mathbb{R}^{N\times(N-1)}$ such that
\begin{equation}
	\begin{pmatrix}
		\frac1{\sqrt N}1_N & U
	\end{pmatrix}
\end{equation}
is orthogonal, and let
\begin{equation}
	\mathcal{U}=
	\begin{pmatrix}
		\frac1N1_N & \frac{1}{\sqrt{N-1}}U
	\end{pmatrix},
\end{equation}
then
\begin{equation}
	\mathcal{U}^{-1}=
	\begin{pmatrix}
		1_N & \sqrt{N-1}U
	\end{pmatrix}^\trans ,
\end{equation}

\subsubsection{Forecast step}
\begin{itemize}
	\item Forward propagation:
	      \begin{equation}
		      x_k^{f,j}=\mathcal{M}_k\left(x_{k-1}^{a,j}\right);
	      \end{equation}
	\item obtain the mean and deviation for $x_k^f$
	      \begin{equation}
		      \left(x_k^f,\Delta_x^f\right)=\left(x_k^{f,j}\right)_j\mathcal{U}
	      \end{equation}
	\item eigen decomposition (approximately):
	      \begin{equation}
		      \left(\Delta_x^f\Delta_x^{f\trans}+\Sigma_k^M\right)V_k\approx V_k\Lambda_k,
	      \end{equation}
	      where $V_k\in\mathbb{R}^{n\times(N-1)}$ and $\Lambda_k\in\mathbb{R}^{(N-1)\times(N-1)}$ are the eigenvectors and eigenvalues of $\Delta_x^f\Delta_x^{f\trans}+\Sigma_k^M$;
	\item update the deviation:
	      \begin{equation}
		      \Delta_x^b=V_k\Lambda_k^{1/2}
	      \end{equation}
	\item update the ensembles:
	      \begin{equation}
		      \left(x_k^{b,j}\right)_j=\left(x_k^f,\Delta_x^b\right)\mathcal{U}^{-1}
	      \end{equation}
\end{itemize}

\subsubsection{Analysis step}
\begin{itemize}
	\item Obtain the ensemble mean and deviation matrix from the ensembles:
	      \begin{equation}
		      \left(x_k^b,\Delta_x^b\right)=\left(x_k^{b,j}\right)_j\mathcal{U}
	      \end{equation}
	\item obtain the mean and deviation for background estimates of observations:
	      \begin{equation}
		      \begin{aligned}
			      y_k^{b,j}                     & =\mathcal{H}_k\left(x_k^{b,j}\right) \\
			      \left(y_k^b,\Delta_y^b\right) & =\left(y_k^{b,j}\right)_j\mathcal{U}
		      \end{aligned}
	      \end{equation}
	\item Transforming matrix:
	      \begin{equation}
		      T_k=\left(I_{N-1}+\Delta_y^{b\trans}\left(\Sigma_k^o\right)^{-1}\Delta_y^b\right)^{-1}=\left(I_{N-1} + S_k^\trans S_k\right)^{-1}
	      \end{equation}
	      for $S_k=\left(\Sigma_k^o\right)^{-1/2}\Delta_y^b$;

	\item normalized innovation vector
	      \begin{equation}
		      \delta_k = \left(\Sigma_k^o\right)^{-1/2}\left(y_k^o-y_k^b\right)
	      \end{equation}
	\item Update the analyzed ensembles: (modified from ETKF)
	      \begin{equation}
		      \begin{aligned}
			      w_k^a      & = \left(I_{N-1} + \Delta_y^{b\trans}\Sigma_k^o \Delta_y^b\right)^{-1}\Delta_y^{b\trans}\left(\Sigma_k^o\right)^{-1}\left(y_k^o-y_k^b\right)=T_kS_k^\trans \delta_k \\
			      x_k^a      & = x_k^b + \Delta_x^bw_k^a = x_k^b + \Delta_x^bT_kS_k^\trans \delta_k                                                                                               \\
			      \Delta_x^a & = \Delta_x^b\left(I_{N-1}+\Delta_y^{b\trans}\left(\Sigma_k^o\right)^{-1}\Delta_y^b\right)^{-1/2}                                                                   \\
			      x_k^{a,j}  & = \left(x_k^a,\Delta_x^a\right)\mathcal{U}^{-1}
		      \end{aligned}
	      \end{equation}
\end{itemize}

\end{document}